\documentclass[11pt]{article}

\usepackage[final]{acl}

\usepackage{times}
\usepackage{latexsym}

\usepackage[T1]{fontenc}

\usepackage[utf8]{inputenc}

\usepackage{microtype}

\usepackage{inconsolata}

\usepackage{graphicx}

\usepackage{booktabs}
\usepackage{wrapfig}
\usepackage{amsmath}
\usepackage{amssymb}
\usepackage{mathtools}
\usepackage{amsthm}
\usepackage{multirow}
\usepackage{subcaption}
\usepackage[table]{xcolor}

\usepackage{listings}
\lstdefinestyle{prompt}{
  basicstyle=\ttfamily\footnotesize,
  breaklines=true,
  breakatwhitespace=true,
  breakindent=1em,
  columns=fullflexible,
  keepspaces=true,
  frame=single  
}

\usepackage{pifont}
\usepackage{listings}
\title{How Far Does Online RL Generalize in Mobile GUI Agents? A Controlled Study with AndroidWorld-Generalization}

\author{
\parbox{\textwidth}{\centering
\vskip 1.5em
\textbf{Li Gu}$^{1,2*\dag}$ \quad
\textbf{Zihuan Jiang}$^{3*}$ \quad
\textbf{Zhixiang Chi}$^{3}$ \quad
\textbf{Huan Liu}$^{4}$ \quad
\textbf{Ziqiang Wang}$^{1,2}$ \\
\textbf{Yuanhao Yu}$^{4}$ \quad
\textbf{Linqiang Guo}$^{2}$ \quad
\textbf{Yang Wang}$^{1,2}$ \\[0.8em]
{\small\mdseries
$^1$Mila -- Qu\'ebec AI Institute \quad
$^2$Concordia University \quad
$^3$University of Toronto \quad
$^4$McMaster University \\
$^*$Equal contribution \quad
$^\dag$Project lead
}
}
}

\begin{document}
\maketitle
\begin{abstract}

Online reinforcement learning (RL) has advanced mobile GUI agents on AndroidWorld
to over 80\% success, but these systems train and evaluate in the same
environment, and their training tasks are not released for auditing. How far RL
generalizes beyond its training contexts therefore remains unmeasured.
We formalize the mobile environment as a Contextual MDP and instantiate a ladder of three progressively distant context shifts as AndroidWorld-Generalization: Unseen Instance, Unseen Template, and Unseen Application, with separation verified manually. In addition, we provide an open RL training system that makes studies at this scale feasible. Applying the identical RL algorithm to two priors of different origin and competence, UI-TARS-7B-SFT at 23.0\% and Qwen3-VL-8B at 52.6\%, we find that RL improves both in distribution but that gains fall monotonically up the ladder (+17.5/+11.1/+4.2 and +10.3/+7.4/+0), with training performance rising throughout while evaluation performance saturates. The decay is algorithm-agnostic, reproduced under PPO, and RL contributes less as the prior strengthens. Few-shot adaptation in the target application recovers +22.9, motivating test-time adaptation as a complement to further RL training. We release the benchmark, training system, recipes, and model weights.

\end{abstract}

\section{Introduction}


Mobile GUI agents --- vision--language models that operate mobile devices by
observing the screen and emitting clicks and keystrokes \citep{gou2025navigating, liu2025llm} --- have
improved dramatically over the past year, with success rates on AndroidWorld
\citep{rawles2025androidworld} climbing from roughly 30\% to over 80\% \citep{xu2026mobilerl}. Much of
this progress comes from online RL: rather than fitting static demonstrations,
which leave the policy prone to error accumulation \citep{bai2024digirl}, the
policy is optimized from its own multi-step trajectories under reward feedback. By
the leaderboard, the problem looks nearly solved.

\setlength{\textfloatsep}{8pt}
\begin{figure*}[t]
  \centering
  \includegraphics[width=\linewidth]{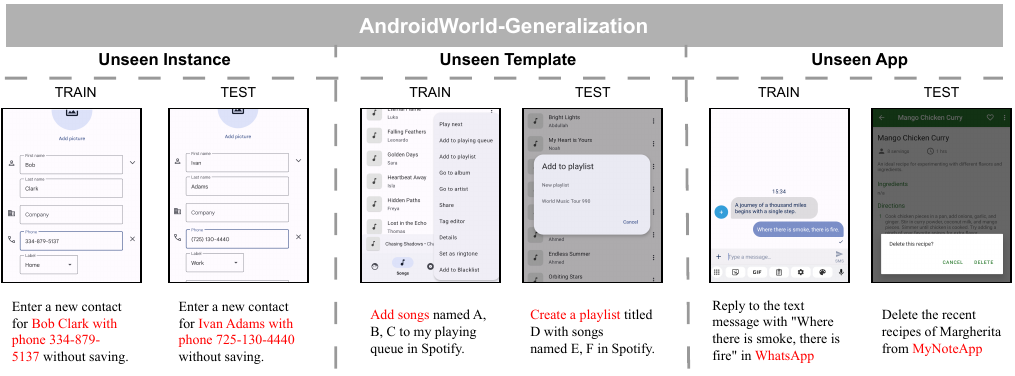}
  \caption{Sample task instructions with corresponding screenshots from the train and test sets of the three unseen regimes in the AndroidWorld-Generalization benchmark. Red highlights the unseen scenarios: Instance, Template, and Application.}
  \label{fig:awg}
\end{figure*}

However, this progress is measured under conditions that differ from
deployment. In practice, mobile agents operate in dynamic, open-ended
environments, where they must handle unseen task instances, unfamiliar
task templates, and entirely new applications. Leaderboard standings do
not speak to this setting. The strongest RL-based methods train inside
the benchmark they evaluate on: MobileRL reaches 80.2\% with a 9B policy
but constructs its training set from 2{,}000 AndroidWorld task instances
\citep{xu2026mobilerl}, and Mobile-Agent-v3 reaches 73.3\% but conducts
its RL stage within the evaluation environment \citep{ye2025mobile}. The
remaining entries conduct no RL in the environment at all, improving
through harness design around frontier models
\citep{gemini3flash2025,seed2026seed1,singh2025openai,liu2026memory}. For agents trained with online RL, how that
generalization degrades as the training--evaluation gap widens remains an open question.

Two obstacles have prevented a direct answer. \textbf{First, no
leakage-controlled measurement protocol exists for online RL in mobile
environments.} Existing benchmarks provide evaluation tasks only, and prior work
that constructs its own training tasks either reuses the evaluation set
\citep{papoudakis2025appvlm} or synthesizes tasks without verifying train--test
separation \citep{shi2025mobilegui, yang2025zerogui}. Because online RL induces
its own training distribution, separation must be enforced over task
specifications rather than held-out episodes, and structured into regimes that
localize where generalization fails. \textbf{Second, no released system provides
the training stack such a protocol requires.} Mobile environments are computationally expensive, delay-prone and
crash-sensitive. The systems reporting the strongest results release model weights
and evaluation code but not the RL training system \citep{xu2026mobilerl,
ye2025mobile}, and the trainers that have been released are configured for other
environments \citep{yang2025zerogui}. Measuring generalization under a controlled
protocol is therefore bounded by the availability of a scalable RL training system
for mobile environments.

In this work, we remove both obstacles. \emph{First,} we formalize the mobile
environment as a Contextual MDP \citep{hallak2015contextual} in which a context
is a task specification, and define a ladder of three context shifts of
increasing semantic distance: Unseen Instance (parameter-level), Unseen Template
(instruction-level), and Unseen Application (domain-level). We instantiate this
ladder as \emph{AndroidWorld-Generalization}, a benchmark for measuring
generalization under online RL. AndroidWorld \citep{rawles2025androidworld}
supplies an evaluation set only, one instance per template; we adopt it as the
base environment for its rule-based reward scripts and automatic task
parameterization, and use that parameterization to construct the training
instances the protocol requires, holding out each regime with separation verified
manually (Fig.~\ref{fig:awg}). \emph{Second,} we develop an open RL training and evaluation system for mobile
environments. Containerized emulators provide resource isolation and crash
recovery, and asynchronous rollout collection removes the synchronization barrier
that otherwise dominates wall-clock time under highly variable trajectory length,
yielding a $6.83\times$ speedup over sequential collection. We release the
training code together with the prompt templates, agent logic, recipes, splits
and model weights, so that training and evaluation are reproducible under a
single protocol.

To enable a controlled study, we apply an identical RL algorithm to two priors of
different origin and initial competence: UI-TARS-7B-SFT \citep{qin2025ui}, a
GUI-specialized model post-trained on agent trajectories, and Qwen3-VL-8B
\citep{bai2025qwen3}, a general-purpose VLM. Both are trained with GRPO
\citep{guo2025deepseek}, and the comparison is replicated under PPO
\citep{schulman2017proximal} on UI-TARS-7B-SFT: a pattern that holds across
independently developed priors cannot be attributed to the particulars of either
training pipeline. Five findings follow,
each answering a question the previous one raises. (1) Online RL improves both
priors in distribution, from 23.0\% to 49.1\% and from 52.6\% to 58.3\%. (2) The
gains do not survive context shift: under a fixed budget and identical
configuration, they fall monotonically up the ladder, $+17.5/+11.1/+4.2$ for
UI-TARS-7B-SFT and $+10.3/+7.4/+0$ for Qwen3-VL-8B, with training
performance rising throughout while evaluation performance saturates, the earlier
the more distant the regime. (3) The decay is algorithm-agnostic: PPO reproduces
the ordering on UI-TARS-7B-SFT ($+17.1/+15.7/+8.3$). (4) RL contributes less as
the prior strengthens, adding $+17.5$ from the 18.8\% prior against $+10.3$ from the
51.7\% prior under the same regime and recipe. (5) The application-level gap is partially recoverable: brief few-shot adaptation
in the target application returns $+22.9$ on UI-TARS-7B-SFT, motivating test-time
adaptation as a complement to further RL training.
\section{Related Works}

\paragraph{Generalization protocols for GUI agents.} Held-out evaluation is
established practice in offline GUI learning: AndroidControl reports unseen-task
and unseen-category splits \citep{li2024effects}, and GUI Odyssey holds
out applications \citep{11444925}. These protocols partition a fixed
corpus of demonstrations, which is sufficient when the training distribution is
fixed by the dataset. Under online RL it is not: the policy generates its own
training data, so separation must be enforced over the task specifications from
which rollouts are drawn. The environments used for online mobile RL supply
evaluation tasks only \citep{rawles2025androidworld, xu-etal-2025-androidlab, kong-etal-2026-mobileworld}. Existing works that
construct training tasks within these environments either reuse the evaluation
set \citep{papoudakis2025appvlm} or generate tasks without verifying
non-overlap \citep{shi2025mobilegui, yang2025zerogui}. In mobile environments,
\citet{lee2025bmoca} vary rendering factors such as icon placement, wallpaper
and device type, holding task semantics fixed and applying no RL training.

\paragraph{Online RL for mobile agents.} Prompting pipelines orchestrate
proprietary VLMs into fixed decision procedures \citep{wen2024autodroid,
li2024appagent, wang2024mobile, agashe2025agent}, and offline post-training fits
human or synthetic trajectories \citep{qin2025ui, wu2025osatlas,
sun-etal-2025-os, cheng2026openmobile}, but static data restricts supervision to per-step accuracy and
cannot capture trajectory-level success \citep{pan2024autonomous}. Online RL
optimizes the policy from its own rollouts, typically by extending GRPO or PPO
with trajectory-level advantages and shaped rewards \citep{bai2024digirl,
shi2025mobilegui, gu2025mobile, ye2025mobile, yang2025zerogui,
tang2026clawgui}, with the strongest reported result on AndroidWorld exceeding
80\% \citep{xu2026mobilerl}. What these systems release, however, differs sharply
from what they report. MobileRL and Mobile-Agent-v3 provide evaluation harnesses
and model weights without a trainer \citep{xu2026mobilerl, ye2025mobile}. ZeroGUI
and ClawGUI release complete trainers, ZeroGUI shipping a desktop-configured pipeline \citep{yang2025zerogui} and ClawGUI
training inside MobileWorld \citep{tang2026clawgui}, but neither defines held-out
context regimes. We release the trainer and the splits together, making
controlled measurement of generalization reproducible.

\paragraph{Generalization in RL.} Generalization is commonly formalized as a CMDP
\citep{hallak2015contextual} with disjoint training and test context sets
\citep{kirk2023survey}, instantiated in classical benchmarks by procedural
generation \citep{cobbe2019quantifying, cobbe2020leveraging} or by varying
environment parameters \citep{packer2018assessing}. For LLM agents the picture
remains unsettled: \citet{chu2025sft} report that RL generalizes where SFT
memorizes, \citet{liu2026paying} that out-of-domain performance can degrade during
RL depending on environment properties and training choices, and
\citet{xi2026can} that generalization is strong across task difficulty but weak
across whole environments. These conclusions are not reconcilable from the
evidence available, and all are obtained in text or grid-world settings such as
WebShop \citep{yao2022webshop}, ALFWorld \citep{shridhar2020alfworld} and Sokoban \citep{junghanns2001sokoban}. We study the same question in a setting none of
these cover: a VLM policy acting on screen observations in a real mobile
environment, with difficulty balanced across splits so that only context distance
varies.

\begin{table*}
\centering
\small
\renewcommand{\arraystretch}{1.2}
\begin{tabular}{lccccccc}
\hline
\textbf{Regime} & \multicolumn{3}{c}{\textbf{Overlap}} & \multicolumn{2}{c}{\textbf{Train}} & \multicolumn{2}{c}{\textbf{Test}}  \\ \cline{2-8} 
                & Instance & Template & App & I/T/A & Diff. & I/T/A & Diff. \\ \hline
Unseen Instance & \ding{55} & \ding{51} & \ding{51} & \textbf{1149} / \textcolor{gray}{78} / \textcolor{gray}{17}        & 1.68       & \textbf{234} / \textcolor{gray}{78} / \textcolor{gray}{17}         & 1.68       \\
Unseen Template & \ding{55} & \ding{55} & \ding{51} & \textbf{836} / \textcolor{gray}{57} / \textcolor{gray}{14}       & 1.70       & \textbf{54} / \textcolor{gray}{18} / \textcolor{gray}{14}        & 1.72       \\
Unseen App      & \ding{55} & \ding{55} & \ding{55} & \textbf{905} / \textcolor{gray}{62} / \textcolor{gray}{12}      & 1.68       & \textbf{48} / \textcolor{gray}{16} / \textcolor{gray}{5}        & 1.69       \\ \hline
\end{tabular}

\caption{\textbf{Statistics of the AndroidWorld-Generalization benchmark}. 
``I", ``T", and ``A" denote Instances, Templates, and Applications; ``Diff." denotes average difficulty. \textbf{Bold numbers }indicate the number of train/test examples in each regime, while \textcolor{gray}{gray numbers} indicate the templates and applications from which those examples are generated.
}
\label{tab:benchmark}
\end{table*}

\section{Benchmarking Generalization in Mobile Environments}

We formalize mobile interactions as a Contextual Markov Decision Process (CMDP)
and introduce AndroidWorld-Generalization, a benchmark that instantiates three
context shifts of increasing semantic distance under strict train--test
separation.

\paragraph{Preliminaries.} We formalize mobile interaction as an MDP
$\mathcal{M} = \langle \mathcal{S}, \mathcal{A}, \mathcal{T}, \mathcal{R}
\rangle$. States $s_t = (o_t, h_t)$ combine the current screenshot $o_t$ with the
interaction history $h_t = \{o_0, a_0, \ldots, o_{t-1}, a_{t-1}\}$; the action
space $\mathcal{A}$ comprises clicking, swiping and typing; and the transition
function $\mathcal{T}$ is determined by the mobile operating system. The policy is
a VLM, $a_t \sim \pi_\theta(\cdot \mid s_t, q)$, conditioned on a natural-language
instruction $q$. Rewards are terminal-only: $r_t = 0$ at every step except a
binary $r_T \in \{0,1\}$ at termination.

\subsection{Generalization in Contextual MDP}

A standard MDP assumes a single stationary environment and cannot express
variation across tasks. We therefore adopt the \emph{Contextual MDP}
\citep{hallak2015contextual}, which factors the state space as $\mathcal{S} =
\mathcal{S}' \times \mathcal{C}$, where $\mathcal{S}'$ is the underlying state
space and $\mathcal{C}$ a context space. A context $c \in \mathcal{C}$ captures
higher-level variation: different instructions within a template, different
templates within an application, or different applications entirely. Before each
episode a context $c \sim P_{\mathcal{C}}$ is sampled and held fixed until the
task terminates, influencing both states and transitions.

To evaluate generalization we adopt \emph{zero-shot policy transfer}
\citep{kirk2023survey}: the context space is partitioned into disjoint subsets
$\mathcal{C}_{\text{train}}$ and $\mathcal{C}_{\text{test}}$ drawn from the same
distribution $P_{\mathcal{C}}$, with $\mathcal{C}_{\text{train}} \cap
\mathcal{C}_{\text{test}} = \emptyset$. The agent trains only on
$\mathcal{C}_{\text{train}}$, and with terminal-only rewards the objective is
\[
\max_{\pi_\theta} \;
\mathbb{E}_{c \in \mathcal{C}_{\text{test}}}\!\left[ r_T \mid c \right],
\]
evaluated without any further training on $\mathcal{C}_{\text{test}}$.

\subsection{AndroidWorld-Generalization}

\noindent \textbf{Why AndroidWorld?} Instantiating this formulation requires an
environment in which held-out contexts can be constructed, not merely evaluated.
AndroidWorld provides two properties that make this possible. First, rule-based
scripts give reliable rewards rather than an LLM judge (\autoref{llm_judge}).
Second, automatic task parameterization generates thousands of distinct instances
from 116 templates spanning three difficulty levels across 20 applications. Newer benchmarks such as MobileWorld \citep{kong-etal-2026-mobileworld} specify each task
with a fixed instruction, which admits no parameter-level shift.

Parameterization induces a natural hierarchy: each application contains multiple
templates, and each template yields many instances under different random seeds.
In the Markor note-taking application, for example, the template
``\textit{Create a new note named \{file\_name\} with the following text:
\{text\}}'' has two parameters, and varying them across seeds produces distinct
instances. AndroidWorld itself ships one instance per template as an evaluation
set; the training distribution used here does not exist in the original benchmark
and is generated through this mechanism.

\noindent \textbf{Train--evaluation task generation.} We treat instances,
templates and applications as three notions of context, and construct one
independent train--test split per regime; each regime therefore has its own
training set, disjoint from its own evaluation set at the corresponding level.
The regimes differ in what they hold constant. \textbf{Unseen Instance} shares
applications and templates across the split and varies only the parameters, so
train and test differ solely in task instantiation. \textbf{Unseen Template}
shares applications but holds out templates, so the agent meets familiar
applications under instructions it has not been trained on; for instance, training
on \textit{create event} and testing on \textit{delete event} within Calendar.
\textbf{Unseen App} holds out applications entirely, so applications, templates
and instances are all disjoint. In every regime, instances are generated with
non-overlapping seed sets.

Two controls make the comparison across regimes interpretable. We match mean
template difficulty between training and evaluation in every regime, within $0.02$
on AndroidWorld's three-level scale (\autoref{tab:benchmark}), so that movement up
the ladder reflects context distance rather than task hardness. And we manually
verify, application by application, that no instance appears on both sides of a
split. Test-set sizes are bounded by the parameterization itself: beyond roughly
sixteen training seeds per template, generated instances begin to duplicate, so
$16/3$ is the largest configuration with guaranteed separation. Split
construction, difficulty computation and the differences from the original
AndroidWorld are detailed in \autoref{awg_benchmark}.

\section{Training System for Mobile Agents}

Building on the benchmark, we present the RL training system. Training mobile
agents in realistic environments requires both an effective learning algorithm and
robust infrastructure for rollout collection, described in \S\ref{sec:grpo} and
\S\ref{sec:rollout} respectively.

\subsection{Online Learning with GRPO}\label{sec:grpo}

Mobile agents perform multi-turn decision-making through interaction with the
environment. We apply the same agent design to two models: at each timestep the
policy conditions on the current screenshot and the full interaction history, and
emits a \textit{thought} followed by an \textit{action} \citep{wei2022chain}, with
the history recording all preceding screenshots, thoughts and actions. Holding the
agent design fixed ensures that differences in outcome reflect the prior rather
than the implementation. \autoref{agent_design} gives the prompt template.

We adopt GRPO \citep{guo2025deepseek} as the online RL algorithm. Given a task
instruction $q$, the policy samples a group of $G$ trajectories
$\{\tau_i\}_{i=1}^{G}$, where $\tau_i=(s_{i,0},a_{i,0},\dots,s_{i,T_i},a_{i,T_i})$
spans $T_i$ timesteps, and optimizes the clipped surrogate


\begin{equation}
\begin{split}
\mathcal{L}(\theta) ={}& -\frac{1}{\sum_{i} T_i}\sum_{i=1}^{G}\sum_{t=1}^{T_i}
\min\Big(r_{i,t}(\theta)\,\hat A_{i,t}, \\
&\qquad \operatorname{clip}\big(r_{i,t}(\theta),1-\epsilon,1+\epsilon\big)\,\hat A_{i,t}\Big)
\end{split}
\end{equation}

with importance ratio $r_{i,t}(\theta)=\pi_\theta(a_{i,t}\mid s_{i,t},q) /
\pi_{\theta_{\text{old}}}(a_{i,t}\mid s_{i,t},q)$. Since the VLM emits actions as
token sequences, ratios are computed at the token level and averaged within each
timestep.

The reward is a single binary signal $R(\tau_i)\in\{0,1\}$ at the terminal step,
so no per-step return is available to score intermediate actions. We therefore
normalize at the trajectory level against the group mean $\mu$ and standard
deviation $\sigma$, and broadcast the result uniformly across steps:
\begin{equation}
\hat A_{\tau_i} = \frac{R(\tau_i)-\mu}{\sigma},
\qquad \hat A_{i,t} = \hat A_{\tau_i}, \;\; \forall t .
\end{equation}
Every action in a successful trajectory is thus credited equally, which avoids
imposing a step-level credit assignment the reward does not identify. We use KL
regularization, omit entropy bonuses, and apply no reward shaping beyond the
terminal signal. The system is not specific to GRPO: we train under PPO
\citep{schulman2017proximal}.

\begin{figure*}[t] 
    \centering
    \includegraphics[width=\textwidth]{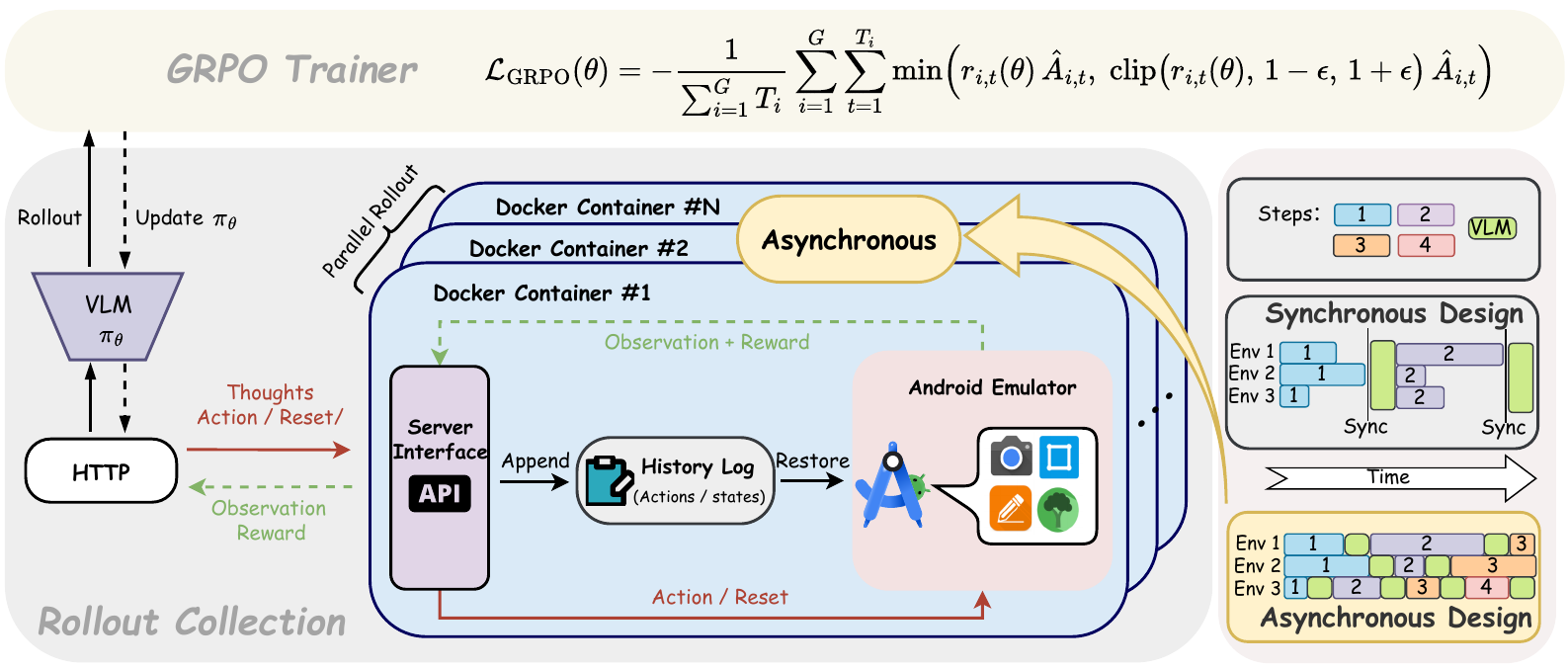}

    \caption{\textbf{RL training system for mobile agent.} We integrate GRPO with a scalable rollout collection system that parallelizes multiple environments. Docker containerization provides resource isolation and decouples trainer and environments through HTTP communication for reliability. Asynchronous rollouts remove the synchronization barrier, enabling more agent steps per unit time. Together, these techniques facilitate reliable and efficient large-scale training.}
    \label{fig:mydiagram}
\end{figure*}

\subsection{Scalable Rollout Collection}\label{sec:rollout}

Rollout collection dominates the cost of each training step
(\autoref{profilling}), so parallel collection is essential: the present study
comprises two models, three regimes and two algorithms with multi-seed evaluation,
which is infeasible to run sequentially. Naive multiprocessing over Android
emulators does not scale, for two reasons. Emulators impose heavy CPU and memory
overhead, so without isolation processes contend for resources and one crashed
environment can terminate collection for all. And because step latency varies
widely across actions, a synchronous implementation waits on the slowest
environment at every step, idling the rest and underutilizing the GPU. Our system
addresses both (\autoref{fig:mydiagram}):

\begin{itemize}\setlength\itemsep{0pt}
\item \textbf{Containerized infrastructure.} Each environment runs in a Docker
container with identical CPU and memory quotas, communicating with the policy over
HTTP. This decouples environment execution from policy updates, so a single
container failure cannot halt training.
\item \textbf{Asynchronous rollouts.} Each environment advances independently and
returns its observation and reward on completion rather than at a global barrier,
removing the synchronization cost and improving GPU utilization.
\end{itemize}


\section{Experiments}

We organize experiments around four questions: Can RL improve mobile agent
performance (Q1)? Does this improvement generalize to unseen contexts (Q2)? Can
few-shot adaptation recover the generalization gap (Q3)? Can the proposed system
accelerate training (Q4)?


\noindent \textbf{Environment and training.} We follow AndroidWorld and use an
Android 13 emulator (API level 33) with 20 pre-installed apps. We train two priors
of different origin and initial competence: UI-TARS-7B-SFT \citep{qin2025ui},
post-trained on agent trajectories (23.0\%), and Qwen3-VL-8B \citep{bai2025qwen3}, a
general-purpose VLM (52.6\%). Each model uses its native agent interface: for
UI-TARS-7B-SFT we adopt its released prompt, and for Qwen3-VL-8B the mobile-use
prompt from the official cookbook, in both cases extended with an \texttt{answer}
action for information-retrieval tasks. Because the Qwen3-VL interface has no
\texttt{open\_app} action and must reach applications through the home screen, we
cap its trajectories at 30 steps against 20 for UI-TARS-7B-SFT. The RL algorithm
and its hyperparameters are otherwise identical across both priors and all three
regimes: eight rollouts per instance, sampling temperature $\tau=1.0$, Adam
\citep{kingma2014adam} at learning rate $1\times10^{-6}$ with 90-step linear
warmup, 16 parallel environments, and four NVIDIA H100-80GB GPUs. All test results
are averaged
over three evaluation seeds per template, with 95\% confidence intervals (\autoref{environment}, \autoref{implementation}).

\noindent \textbf{Baselines.} We compare against proprietary systems,
supervised fine-tuned open models, and recent RL methods at the 3--9B scale. We
separate systems we evaluate ourselves under a controlled protocol from systems
whose training data, task splits, or recipes are unreleased; the latter are
reported as published and are not directly comparable.

\begin{figure}[h]
\centering
\begin{subfigure}{\linewidth}
    \centering
    \includegraphics[width=\linewidth]{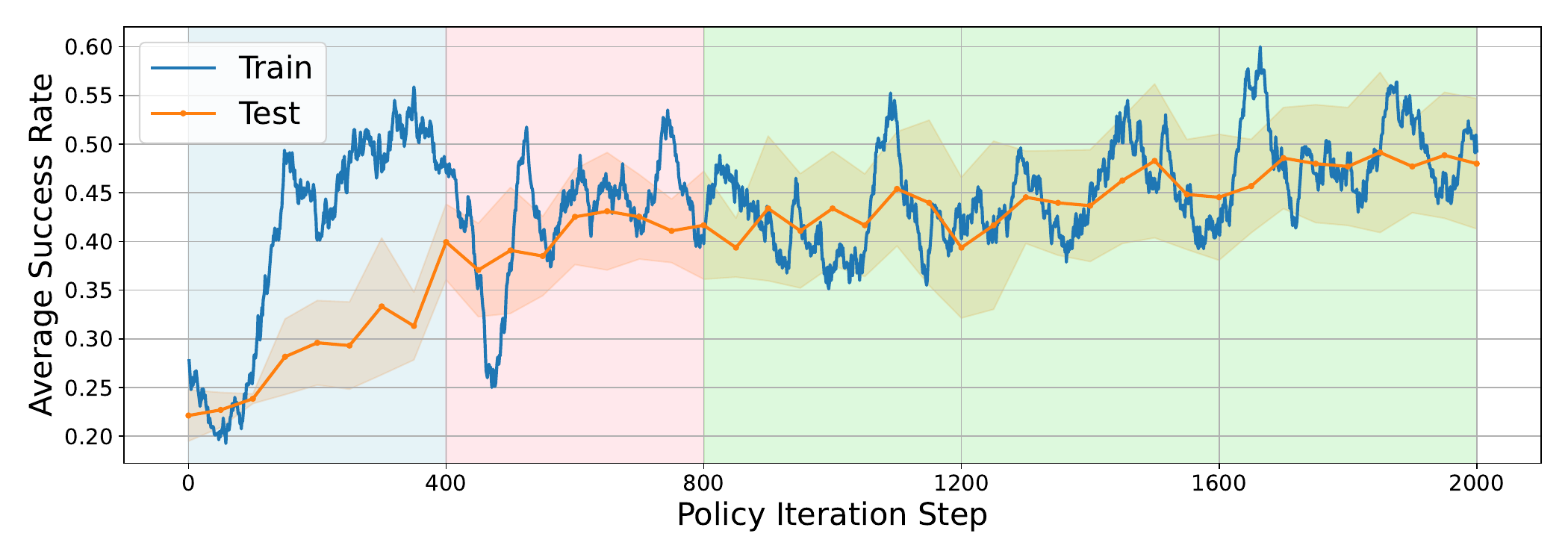}

\end{subfigure}
\hfill
\begin{subfigure}{\linewidth}
    \centering
    \includegraphics[width=\linewidth]{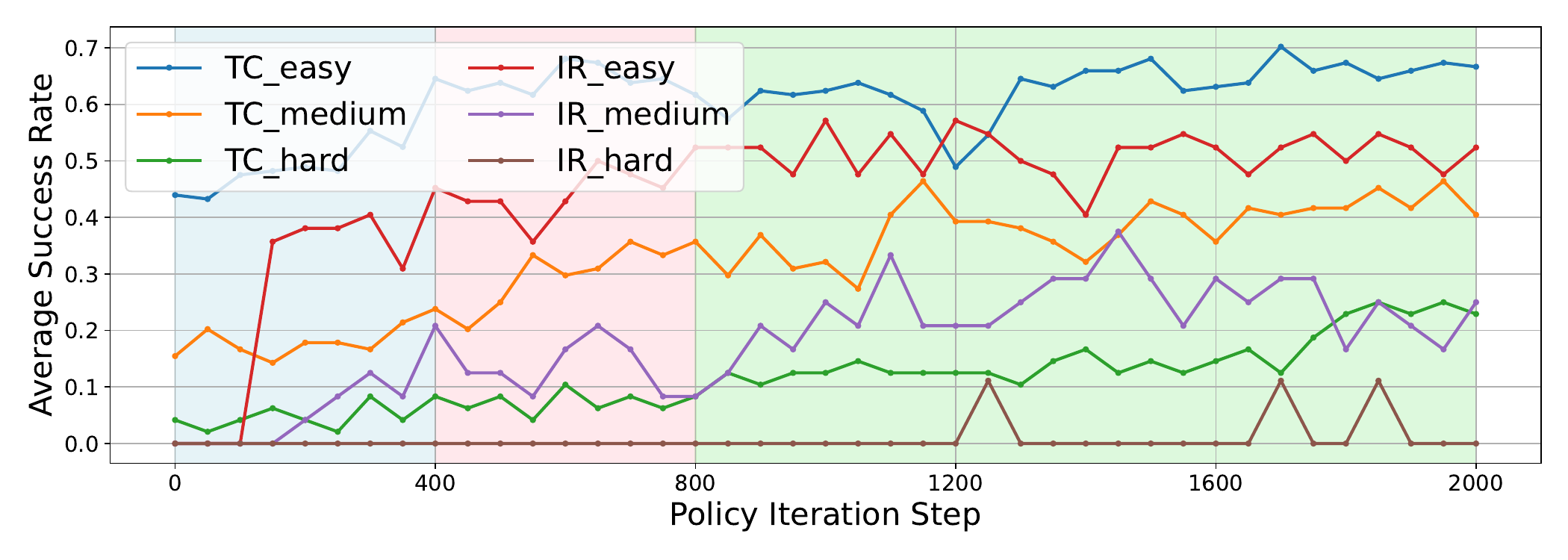}

\end{subfigure}
\vspace{-10pt}
\caption{Training dynamics on Unseen Instance with curriculum learning. Colored areas denote curriculum stages: blue (Easy), red (Easy + Medium), green (All). (Top) Average training and evaluation success rates. (Bottom) Average evaluation success rates by task type (Task Completion, Information Retrieval) and difficulty (Easy, Medium, Hard).}
\label{fig:main_result}
\end{figure}

\textbf{Q1: Can RL improve the decision-making capabilities of mobile agents?}
We first evaluate whether RL post-training improves over the SFT baseline on the
standard AndroidWorld benchmark. Although the Unseen Instance regime trains on 78
templates, we evaluate on all 116 templates for comparability with prior work. 
All test results are obtained by evaluating saved checkpoints after training
completes; no test signal is used for checkpoint selection or hyperparameter
tuning.
Building on UI-TARS-7B-SFT, our RL method employs curriculum learning that expands
training tasks from easy, to easy + medium, and finally to the full task set. This
more than doubles the average success rate, from 23.0 to 49.1 ($+26.1$ points,
\autoref{tab:leaderboard_v2}), with consistent gains across all difficulty
levels---28.4\% on Easy, 26.8\% on Medium, and 17.5\% on Hard
(\autoref{fig:main_result}). Applied to the stronger Qwen3-VL-8B base, RL lifts
the success rate from 52.6 to 58.3 ($+5.7$).
The strongest Panel~B systems report higher absolute scores, but their domain-specific SFT
data curation, splits, and RL recipes are unreleased and build on different base
models. These differences make direct attribution difficult, and we cite them
as published and focus our comparison on the controlled protocol of Panel~A.

The per-category evaluation in \autoref{fig:main_result} (bottom) further
demonstrates gains across task types, particularly for information retrieval which
starts from a success rate of 0. Training on easy tasks transfers to medium and hard
ones, but hard information retrieval remains unsolved.

\begin{table*}[t]
\centering 
\small
\renewcommand{\arraystretch}{1.2}
\begin{tabular*}{\textwidth}{@{\extracolsep{\fill}}lcccc@{}}
\toprule
{\textbf{Models}} & \textbf{API-Based} & \textbf{Open Weights} & \textbf{Recipe \& Splits} & \textbf{Average (SR)} \\
\midrule
\rowcolor{gray!20}\multicolumn{5}{@{}l}{\textbf{Panel A: controlled protocol}} \\
\multicolumn{5}{@{}l}{\textit{Zero-shot, no AndroidWorld training}} \\
Claude Computer Use \citep{anthropic2024claude35} &\ding{51}&&---& 27.9 \\
GPT-5.4 \citep{singh2025openai} &\ding{51}&&---& 56.0 \\
Gemini-3-Flash \citep{gemini3flash2025} &\ding{51}&&---& 50.6 \\
Seed1.8 \citep{seed2026seed1} &\ding{51}&&---& 70.7 \\
UI-TARS-7B-SFT \citep{qin2025ui} &&\ding{51}&---& 23.0 $\pm$ 2.2 \\
Qwen3-VL-8B \citep{bai2025qwen3} &&\ding{51}&---& 52.6 $\pm$ 3.8 \\
\cmidrule(lr){1-5}
\multicolumn{5}{@{}l}{\textit{Trained on AndroidWorld-Generalization (ours)}} \\
Ours (UI-TARS-7B) w/o curriculum &&\ding{51}&\ding{51}& 45.1 $\pm$ 2.5 (\textcolor{green!60!black}{+22.1}) \\
Ours (UI-TARS-7B) &&\ding{51}&\ding{51}& \textbf{49.1 $\pm$ 8.2 (\textcolor{green!60!black}{+26.1})} \\
Ours (Qwen3-VL-8B) w/o curriculum &&\ding{51}&\ding{51}& \textbf{58.3 $\pm$ 3.5 (\textcolor{green!60!black}{+5.7})} \\
\midrule
\rowcolor{gray!20}\multicolumn{5}{@{}l}{\textbf{Panel B: reported as published}} \\
\textcolor{gray}{AppVLM-3B \citep{papoudakis2025appvlm}} &&&& \textcolor{gray}{37.8} \\
\textcolor{gray}{GUI-Owl-7B \citep{ye2025mobile}} &&\ding{51}&& \textcolor{gray}{66.4} \\
\textcolor{gray}{MobileRL-9B \citep{xu2026mobilerl}} &&\ding{51}&& \textcolor{gray}{80.2} \\
\textcolor{gray}{ATMem-UI-8B} \citep{liu2026memory} &&&& \textcolor{gray}{76.6} \\
\bottomrule
\end{tabular*}
\caption{Results on AndroidWorld. \textbf{Panel A}: models with no AndroidWorld training (zero-shot) or models trained by us on released splits. Entries with ± are evaluated by us under our controlled protocol and reported as mean ± standard deviation over three evaluation seeds per template; entries without ± are cited as published. \textbf{Panel B}:
published numbers from systems with unreleased data, splits, or recipes, several
trained inside AndroidWorld; cited as reported. ``---'': not trained on
AndroidWorld. Curriculum used for UI-TARS-7B only. Full table in
\autoref{tab:appendix_b}.}
\label{tab:leaderboard_v2}
\end{table*}

\begin{figure*}[!htb]
\centering
\includegraphics[width=\textwidth]{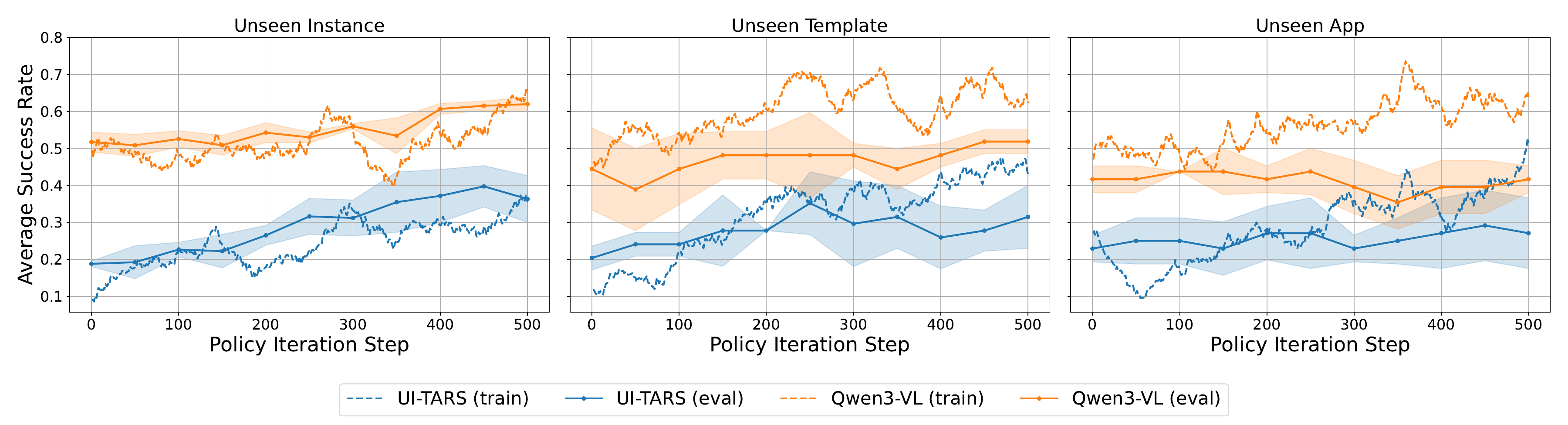}
\caption{Training dynamics of GRPO across the three unseen regimes. We report training success rates and evaluation success rates with standard deviations.}
\label{fig:ablation_study_on_unseen_scenarios}
\end{figure*}

\textbf{Q2: Does the improvement generalize to increasingly distant
contexts?}\label{sec:q2}
We now evaluate generalization across the three regimes of
AndroidWorld-Generalization. We fix a 500-step budget without curriculum and hold every other setting constant
across regimes and priors, isolating the effect of context shift. Gains below are
therefore not comparable to the $+26.1$ of Q1: the $+17.5$ on Unseen Instance is
the same prior measured under a shorter budget on a smaller held-out set.
Difficulty is matched between training and evaluation in every regime
(\autoref{tab:benchmark}), so movement up the ladder reflects context distance
rather than task hardness.

\noindent\textbf{Gains decay with context distance.} As shown in
\autoref{fig:ablation_study_on_unseen_scenarios}, evaluation success rates rise
substantially on Unseen Instance and progressively less thereafter: $+17.5 /
+11.1 / +4.2$ for UI-TARS-7B-SFT and $+10.3 / +7.4 / +0$ for
Qwen3-VL-8B. The two priors differ in magnitude throughout but produce the same
ordering. Given the held-out set sizes ($n = 234/54/48$), we claim this ordering
and the widening train--test gap rather than a separable difference between
individual regimes; confidence intervals are reported in \autoref{appendix:ci}.

\noindent\textbf{Training improves while evaluation saturates.} Training success rises over the full 500 steps in every regime, whereas
evaluation success plateaus, and does so earlier the more distant the regime.
The training--evaluation gap therefore widens as training proceeds, and is
widest under Unseen App, where the context shift is largest.

\noindent\textbf{The decay is algorithm-agnostic.} Training UI-TARS-7B-SFT with
PPO under the same training recipe and budget reproduces the pattern
(${+}17.1/{+}15.7/{+}8.3$; \autoref{fig:ppo_unseen_scenarios}). Both algorithms
rank the three unseen regimes identically.

\noindent\textbf{RL contributes less as the prior strengthens.} On Unseen Instance, the
same recipe lifts the $18.8\%$ prior to $36.3\%$ ($+17.5$) but the $51.7\%$ prior only to
$62.0\%$ ($+10.3$); the same gap holds on Unseen Template and Unseen App. The stronger prior still ends higher in every regime. Because the
two priors differ in architecture, data, and post-training as well as in initial
competence, we describe this as a property of the two systems studied, not a general
relationship.

\noindent\textbf{Case study.} To understand what enables transfer on Unseen
Template, we analyze templates that failed before RL and succeeded after. Although
the template is unseen, completing it requires skills acquired from seen
templates: an unseen template requiring deletion of a recipe from a list relies on
identifying target content, a skill transferred from a seen template involving
food-name search. Further examples appear in \autoref{unseen_template}.

\begin{figure}[h!]
    \centering
    \includegraphics[width=\linewidth]{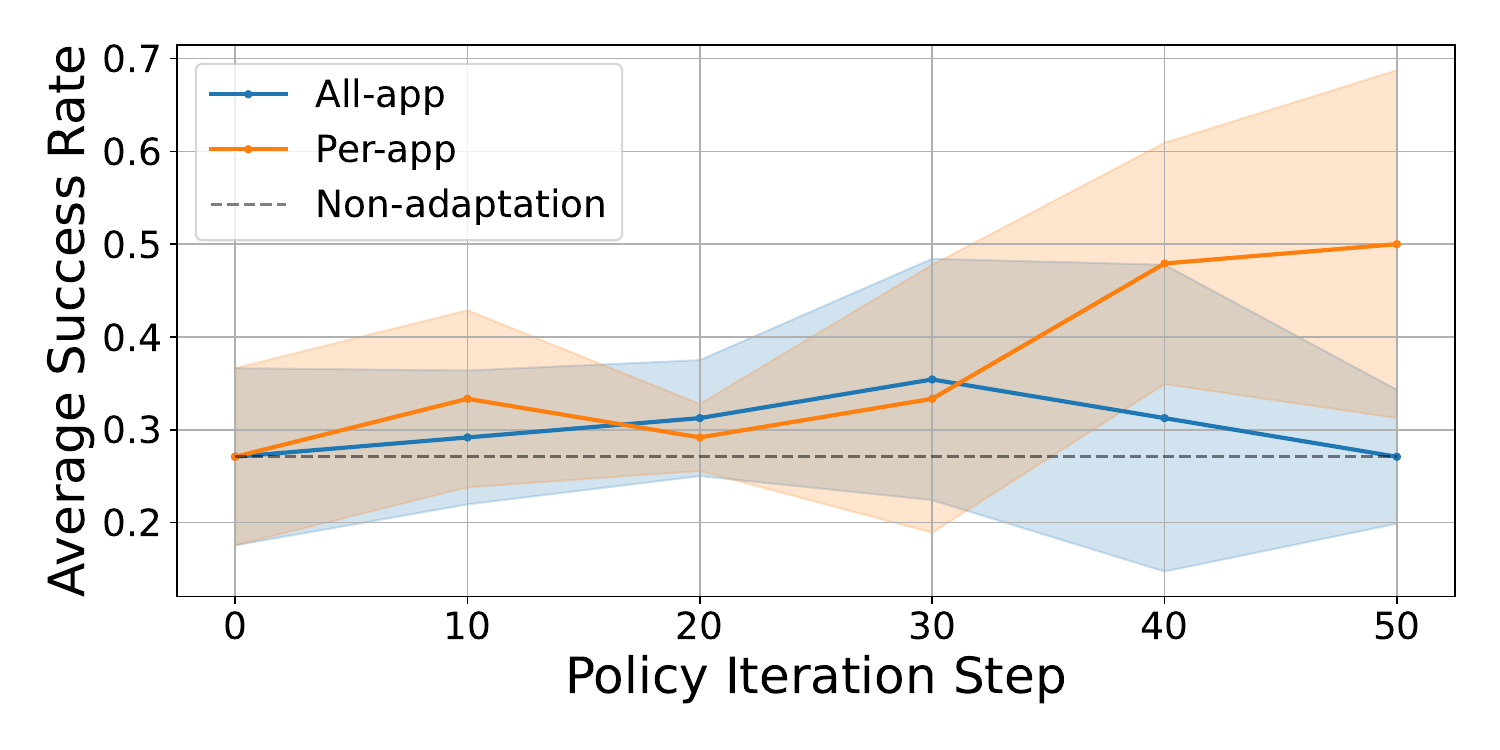}
    \caption{Few-shot adaptation vs. non-adaptation, averaged over Unseen App test-set.}
    \label{fig:unseen_app_tta_v2}
\end{figure}

\textbf{Q3: Can few-shot adaptation at test time recover the
generalization gap?}
If the app-level gap reflects missing exposure to the target context, a small
amount of experience there should partly close it. We test this on
UI-TARS-7B-SFT in the Unseen App regime. Starting from the model trained for 500
steps on seen apps, we allow 50 additional fine-tuning steps on 8 instances per
unseen app, reflecting the limited data and compute available in deployment, and
compare two strategies: \textbf{All-App}, fine-tuning jointly on all unseen-app
instances, and \textbf{Per-App}, fine-tuning a separate model for each
application. 

As shown in \autoref{fig:unseen_app_tta_v2}, Per-App adaptation improves over the
non-adapted baseline by $22.9$ ($27.1\%$ to $50.0\%$) and is still improving when
the budget ends, so this is a lower bound on what 8 target-application instances
can recover. All-App adaptation instead yields a transient gain, peaking at $+8.3$
around step 30 and returning to baseline by step 50. The two strategies differ in
both specialization and total adaptation compute, so we do not attribute the
difference between them to either alone. Because adaptation and test instances
share templates, $22.9$ upper-bounds recovery from application shift alone
(\autoref{appendix:fewshot}).



\textbf{Q4: Can the proposed rollout collection system accelerate RL
training?}
Rollout collection dominates each training step, while policy update time is
unaffected by parallelism. Using 16 environments in parallel reduces collection
time by $6.83\times$ compared to a single environment in the sequential setting.
Asynchrony contributes independently: without it, the trainer waits for every
environment in a group to finish, so the longest rollout bottlenecks progress, an
effect that amplifies with group size and costs a $57.8\%$ slowdown at 16
environments. Together these make the present study feasible, comprising two
priors, three regimes, two algorithms, and multi-seed evaluation at every
checkpoint. Detailed results and settings are given in \autoref{profilling} and
\autoref{rollout_ablate}.


\section{Conclusion}

We study the generalization of online RL for mobile GUI agents when evaluation
contexts differ from training contexts. To answer the question, we formalize the
mobile environment as a Contextual MDP, introduce AndroidWorld-Generalization, a
benchmark instantiating three progressively distant unseen regimes under manually
verified separation, and release an open RL training system. Applying an identical
RL algorithm to two priors of different origin, under both GRPO and PPO, we find a
structured decay: RL yields strong gains on unseen instances, but performance
decays on unseen templates and apps. The pattern holds for both priors and both
algorithms, and RL contributes less as the prior strengthens. Few-shot per-app
adaptation partially recovers the gap. Two future directions follow from these findings: adaptation methods that acquire
target-context coverage efficiently, and supervised stages designed to maximize
what the subsequent RL stage can add.

\clearpage 
\section{Limitations}
Our held-out sets are small ($n=234/54/48$), and the 95\% intervals on Unseen
Template and Unseen App overlap substantially (\autoref{appendix:ci}); we
therefore claim the ordering of regimes and the widening train--test gap rather
than a statistically separable difference between individual regimes. In addition,
the three unseen regimes span 20 applications and 116 task templates, with only
five applications held out in Unseen App, so app-level conclusions rest on a
narrow slice of deployment scenarios. Since pretraining corpora are not fully
disclosed for any open VLM, ``unseen'' here denotes held out from the training we
perform and disclose, not from all prior exposure. Plus, our two priors differ
in architecture, data, and post-training procedure as well as in initial
competence, so the observation that RL contributes less as the prior strengthens
describes the two systems studied rather than a general relationship. Finally, although the rollout collection system is built on
containerized infrastructure designed to scale to hundreds of environments across
multi-node clusters, our experiments run on a single node under the resource constraints
typical of academic settings, so its scalability is argued from design rather than
demonstrated, and training throughput is bounded accordingly.

\section{Impact Statements}
Mobile agents capable of automating GUI interactions have the potential to improve accessibility for users with disabilities and reduce repetitive manual tasks. For instance, such agents could assist visually impaired users in navigating complex mobile interfaces or help elderly users complete routine digital tasks independently. However, such technology could also be misused for malicious purposes, such as unauthorized access to devices, automated spamming, or bypassing security mechanisms. Additionally, widespread deployment of autonomous agents raises concerns about user privacy, as these systems may interact with sensitive personal data during task execution. We encourage future work to incorporate safety mechanisms, access controls, and privacy-preserving designs to mitigate potential misuse.







\bibliography{iclr2026_conference}

\appendix

\section{Reproducibility Statement}
Our proposed benchmark and RL system for mobile use agents are fully open-sourced. Task instances for all three regimes are publicly released, with detailed generation procedures provided in \autoref{awg_benchmark}. We additionally release the complete training framework, including the interactive environment and configurations, task sets, prompt templates, agent logic, training hyperparameters, and RL infrastructure, covering both the trainer and the rollout collection system. The anonymized codebase is available at \url{https://anonymous.4open.science/r/AndroidWorldGeneralization-BE55}.


\section{LLMs Usage}
ChatGPT-5 is used solely as a general-purpose writing assistant. Specifically, we apply a fixed prompt template — “Polish the writing in a concise and academic way” — to improve grammar, clarity, and style of text written by the authors. The LLM did not contribute to research ideation, methodology design, experimental setup, analysis, or result interpretation.

\section{AndroidWorld-Generalization Benchmark}
\label{awg_benchmark}

\paragraph{AndroidWorld. }AndroidWorld is an interactive benchmark built on the Android Emulator, pre-installed with 20 applications and 116 task templates. Each template defines a workflow (e.g., creating a calendar event, deleting a recipe entry) with parameter slots (e.g., event name, date, recipe type) that can be instantiated with different values to generate unlimited task instances via random seeds. Tasks span two categories: \textit{task completion} (e.g., opening an app and completing a form) and \textit{information retrieval} (e.g., extracting a contact name from the address book). Templates are further grouped into three difficulty levels—\textit{easy}, \textit{medium}, and \textit{hard}—based on the number of steps and reasoning complexity required. Task success is evaluated via rule-based scripts. However, AndroidWorld was originally designed for evaluation rather than training, and its codebase only supports sequential execution without parallel rollout collection.

\paragraph{Train-Test Split on Unseen Instance} 
For evaluation, we follow the AndroidWorld protocol and use all 116 task templates, enabling direct comparison with prior work. 
For each template, we report mean and standard deviation over three evaluation seeds---seed 30 (the AndroidWorld default) and seeds 7 and 1234---chosen to maximize the number of unique instances. 
For training, we exclude 38 templates that consistently produce identical task instances regardless of seed, leaving 78 templates. 
From these, we generate distinct task instances using 16 random seeds 
\{1, 2, 3, 4, 5, 6, 8, 9, 12, 123, 12345, 123456, 1234567, 12345678, 123456789, 1234567890\}, 
yielding $16 \times 78 = 1248$ instances. 
To avoid overlap between train and test, we manually remove duplicates, resulting in a final training set of 1149 unique task instances.

\paragraph{Train-Test Split on Unseen Template}
Given the 116 templates in AndroidWorld, we construct the Unseen Template scheme by splitting them into training and evaluation sets according to the following rules. First, we require both seen and unseen templates in the training and evaluation sets to come from overlapping applications. Therefore, we filter out applications that contain only a single template. To further divide training and evaluation, we target a 3:1 ratio while maintaining comparable average difficulty levels (difficulty is defined in the following subsection). Specifically, for each application, if a medium- or hard-level template is assigned to the evaluation set, at least one template of equal or lower difficulty from the same application is retained in the training set. For each template, we then generate distinct task instances using the same automatic generation mechanism with non-overlapping random seeds, as described earlier.

As a result, the filtered dataset contains 14 applications: the training set includes 57 templates (mean difficulty = 1.70) with 836 instances, and the evaluation set includes 18 templates (mean difficulty = 1.72) with 54 instances.

\paragraph{Train-Test Split on Unseen App}  
Following the Unseen Instance scheme, 38 templates could not be used for instance generation due to random seed constraints in three applications, leaving 17 applications for the Unseen App regime. Applying the same 3:1 split rule and ensuring similar difficulty levels, we construct the dataset as follows: the training set consists of 62 templates (mean difficulty = 1.68) with 905 instances from 12 applications, while the evaluation set consists of 16 templates (mean difficulty = 1.69) with 48 instances from 5 non-overlapping applications.


\paragraph{Difficulty calculation}
AndroidWorld includes 116 task templates, each assigned a difficulty level: easy (1), medium (2), or hard (3). To maintain comparable difficulty distributions between training and testing splits across the three unseen regimes (Unseen Instance, Unseen Template, and Unseen App), we ensure that the ratio of templates across difficulty levels is preserved. For example, in the Unseen Template regime, easy, medium, and hard templates are evenly divided between training and testing. Similarly, in the Unseen App regime, we select app combinations such that the proportions of easy, medium, and hard templates in the training apps match those in the held-out test apps. As a result, the average difficulty level between training and testing splits remains similar across all regimes, ensuring that generalization is evaluated under comparable task complexity, reported in \autoref{tab:benchmark}.

\begin{figure*}[h!]
\centering
\begin{minipage}[]{0.48\linewidth}
    \centering
    \includegraphics[width=\linewidth]{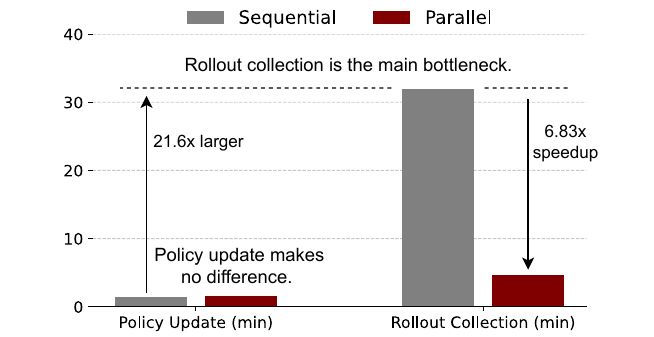}
\end{minipage}
\begin{minipage}[]{0.48\linewidth}
    \centering
    \includegraphics[width=\linewidth]{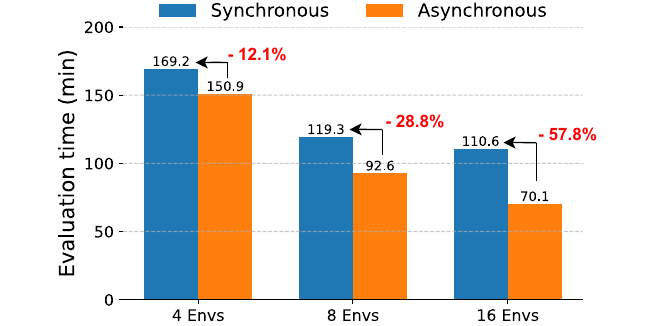}
\end{minipage}
\caption{(Left) Time profiling of a policy update vs. the collection of 16 rollouts per training step. (Right) Performance of async vs. sync rollouts with varying environment counts.}
\label{profilling}
\end{figure*}

\paragraph{AndroidWorld-Generalization vs.\ AndroidWorld}
A standard benchmark for agentic mobile tasks typically consists of three components: an interactive environment, a task suite with an associated verifier, and a codebase capable of generating and evaluating rollouts. Along the environment dimension, AndroidWorld and AndroidWorld-Generalization are identical: both use the same 20 Android applications and 116 manually curated task templates. However, the remaining dimensions differ substantially. First, in terms of the task suite, AndroidWorld provides only a single task instance per template (116 total) for evaluation, which limits any robustness assessment. In contrast, AndroidWorld-Generalization uses the automatic task-parameterization mechanism to generate three evaluation instances per template, and leverages the remaining non-overlapping seeds to construct thousands of additional instances that form standardized training sets across three unseen generalization regimes (instance, template, and application). Second, regarding RL support, AndroidWorld was designed solely for evaluation (released in May 2024), before the community’s shift toward LLM-driven RL (e.g., DeepSeek-R1, Jan.\ 2025), and therefore includes no RL-capable training split or interface. AndroidWorld-Generalization explicitly supports the RL paradigm, enabling reproducible RL training and systematic study of RL generalization. Third, with respect to infrastructure, AndroidWorld’s evaluation pipeline is fully sequential, making large-scale testing prohibitively slow. AndroidWorld-Generalization introduces a parallelized rollout engine (Section~4), providing up to a 6.83$\times$ speedup (\autoref{profilling}) and supporting scalable RL training and evaluation. A comparison is summarized in Table~\ref{tab:awg-vs-aw}.

\begin{table*}[h!]

\centering
\vspace{4pt}
\begin{tabular}{p{3.4cm} p{4.1cm} p{5.8cm}}
\toprule
 & \textbf{AndroidWorld} & \textbf{AndroidWorld-Generalization} \\
\midrule
\textbf{Environment} &
116 templates across 20 apps &
Same 20 apps and templates (subset splits) \\
\addlinespace[4pt]

\textbf{Task Suite} &
116 eval instances (1 seed / template) &
Multi-seed eval instances; thousands of non-overlapping train instances \\
\addlinespace[4pt]

\textbf{Generalization} &
None &
Three unseen regimes (instance, template, app) \\
\addlinespace[4pt]

\textbf{RL Support} &
No; evaluation-only &
Yes; standardized train set for RL \\
\addlinespace[4pt]

\textbf{Infrastructure} &
Sequential rollouts &
Parallel rollout collection for scalable evaluation \& RL training \\
\bottomrule
\end{tabular}
\caption{Comparison between AndroidWorld and AndroidWorld-Generalization.}
\label{tab:awg-vs-aw}
\end{table*}

\section{Additional experiment results}
\label{PPO_exp_result}
\begin{figure*}[h!]
\centering
\includegraphics[width=\textwidth]{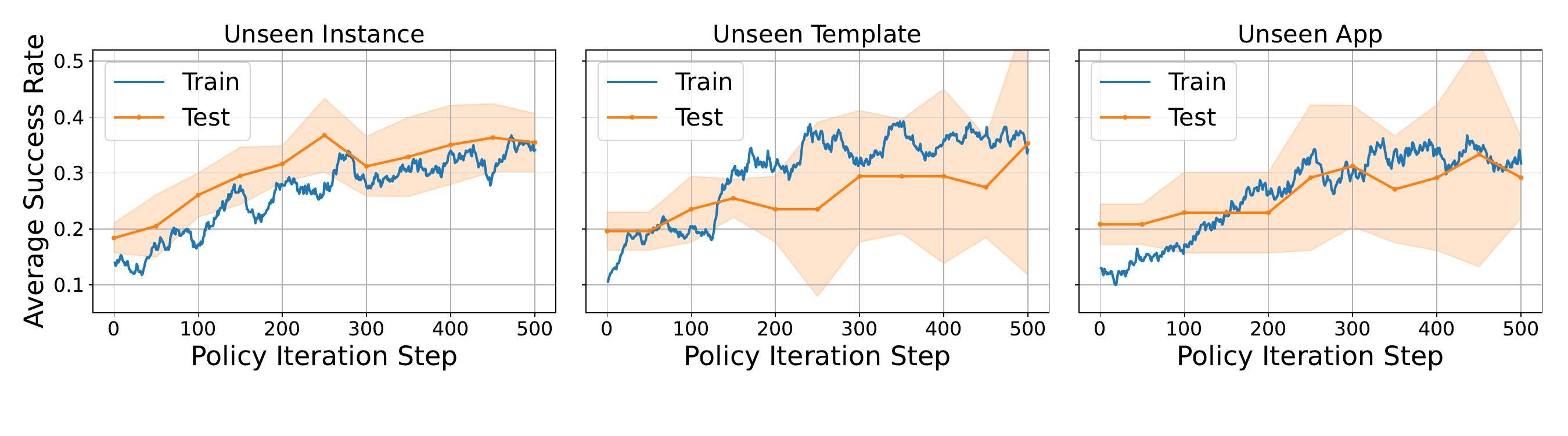}
\vspace{-30pt}
\caption{Training dynamics of PPO across the three unseen regimes. We report training success rates and evaluation success rates with standard deviations.}
\label{fig:ppo_unseen_scenarios}
\end{figure*}

\paragraph{PPO} To demonstrate that our RL training system is algorithm-agnostic, we also evaluate PPO on all three unseen regimes in AndroidWorld-Generalization. We follow a simple PPO baseline with token-level GAE as in \citep{wang2025vagen}, where a binary reward is assigned only at the final token of each trajectory and propagated backward to compute token-level advantages. As shown in \autoref{fig:ppo_unseen_scenarios}, PPO exhibits the same qualitative trends as GRPO: evaluation success rates improve by 17.1\% on Unseen Instance, 15.7\% on Unseen Template, and 8.3\% on Unseen App.

\label{unseen_instance}


\begin{table*}[h!]
\centering
\small
\resizebox{\textwidth}{!}{
\begin{tabular}{lcccc}
\toprule
{\textbf{Models}} & \textbf{Easy (SR)} & \textbf{Medium (SR)} & \textbf{Hard (SR)} & \textbf{Average (SR)} \\
\midrule
\rowcolor{gray!20}\textit{Closed-source Models} \\
GPT-4o \citep{hurst2024gpt} &-&-&-& 34.5 \\
GPT-5.4 \citep{singh2025openai}  &-&-&-& 56.0 \\
Claude Computer Use \citep{anthropic2024claude35} &-&-&-& 27.9 \\
Gemini-3-Flash \citep{gemini3flash2025}  &-&-&-& 50.6 \\
Seed1.8 \citep{seed2026seed1}  &-&-&-& 70.7 \\

\midrule
\rowcolor{gray!20}\textit{Open-source 32B/72B Models} \\
Qwen2.5-VL-32B \citep{bai2025qwen2} &-&-&-& 31.5 \\
MobileGUI-32B \citep{shi2025mobilegui} &-&-&-& 44.8 \\
AGUVIS-72B \citep{xu2024aguvis} &-&-&-& 26.1 \\
Qwen2.5-VL-72B \citep{bai2025qwen2} &-&-&-& 35.0 \\
UI-TARS-72B-SFT \citep{qin2025ui} &-&-&-& 46.6 \\
MobileUse-72B \citep{li2025mobileuse} & 83.6 & 47.2 & 26.3 & 62.9 \\

\midrule
\rowcolor{gray!20}\textit{Open-source 2B/7B/8B Models}\\
AppVLM-3B\textsuperscript{*} \citep{papoudakis2025appvlm} & 57.9 & 27.4 & 8.3 & 37.8 \\
MobileGUI-7B \citep{shi2025mobilegui} &-&-&-& 30.0 \\
GUI-Owl-7B \citep{ye2025mobile} &-&-&-& 66.4 \\
MAI-UI-8B \citep{zhou2025mai} &-&-&-& 70.7 \\
MobileRL-7B \citep{xu2026mobilerl} &-&-&-& 72.0 \\
MobileRL-9B \citep{xu2026mobilerl} &-&-&-& 80.2 \\
ATMem-UI-8B \citep{liu2026memory} &-&-&-& 76.6 \\

\cmidrule(lr){1-5}

UI-TARS-7B-SFT \citep{qin2025ui} & 33.9 ± 5.3 & 13.9 ± 2.8 & 5.3 ± 0.0 & 23.0 ± 2.2 \\
Ours (UI-TARS-7B) w/o curriculum learning & 62.8 ± 0.9 & 31.5 ± 8.0 & 14.0 ± 3.0 & \textbf{45.1 ± 2.5 (\textcolor{green!60!black}{+22.1})} \\
Ours (UI-TARS-7B) & 62.3 ± 5.7 & 40.7 ± 11.6 & 22.8 ± 13.2 & \textbf{49.1 ± 8.2  (\textcolor{green!60!black}{+26.1})} \\

Qwen3-VL-8B \citep{bai2025qwen3} & 67.2 ± 3.3 & 46.3 ± 5.8 & 17.5 ± 6.1 & 52.6 ± 3.8 \\
Ours (Qwen3-VL-8B) w/o curriculum  & 71.6 ± 4.1 & 54.6 ± 5.8 & 22.8 ± 3.0 & \textbf{58.3 ± 3.5  (\textcolor{green!60!black}{+5.7})} \\
\bottomrule
\end{tabular}
}
\caption{AndroidWorld evaluation performance for Mobile Agents.\protect\footnotemark}
\label{tab:appendix_b}
\end{table*}
\footnotetext{Prior methods marked with $^{*}$ are evaluated on a subset of AndroidWorld.}

\section{Case study on Unseen Template}
\label{unseen_template}
To investigate the underlying reasons for zero-shot transfer in the \textbf{Unseen Template} regime, we analyze task instances that succeed after RL training but fail before. Although task templates are disjoint, completing a template typically requires one or a few fundamental skills, which can transfer from seen to unseen templates. This skill sharing explains why policies trained on certain templates can generalize to non-overlapping ones. We highlight task instances from all five unseen templates that achieve success after RL training, identify the transferable skills they require, and trace these skills back to corresponding seen templates that provided them during training, as summarized below.

\begin{figure*}[h!]
\centering
\includegraphics[width=\textwidth]{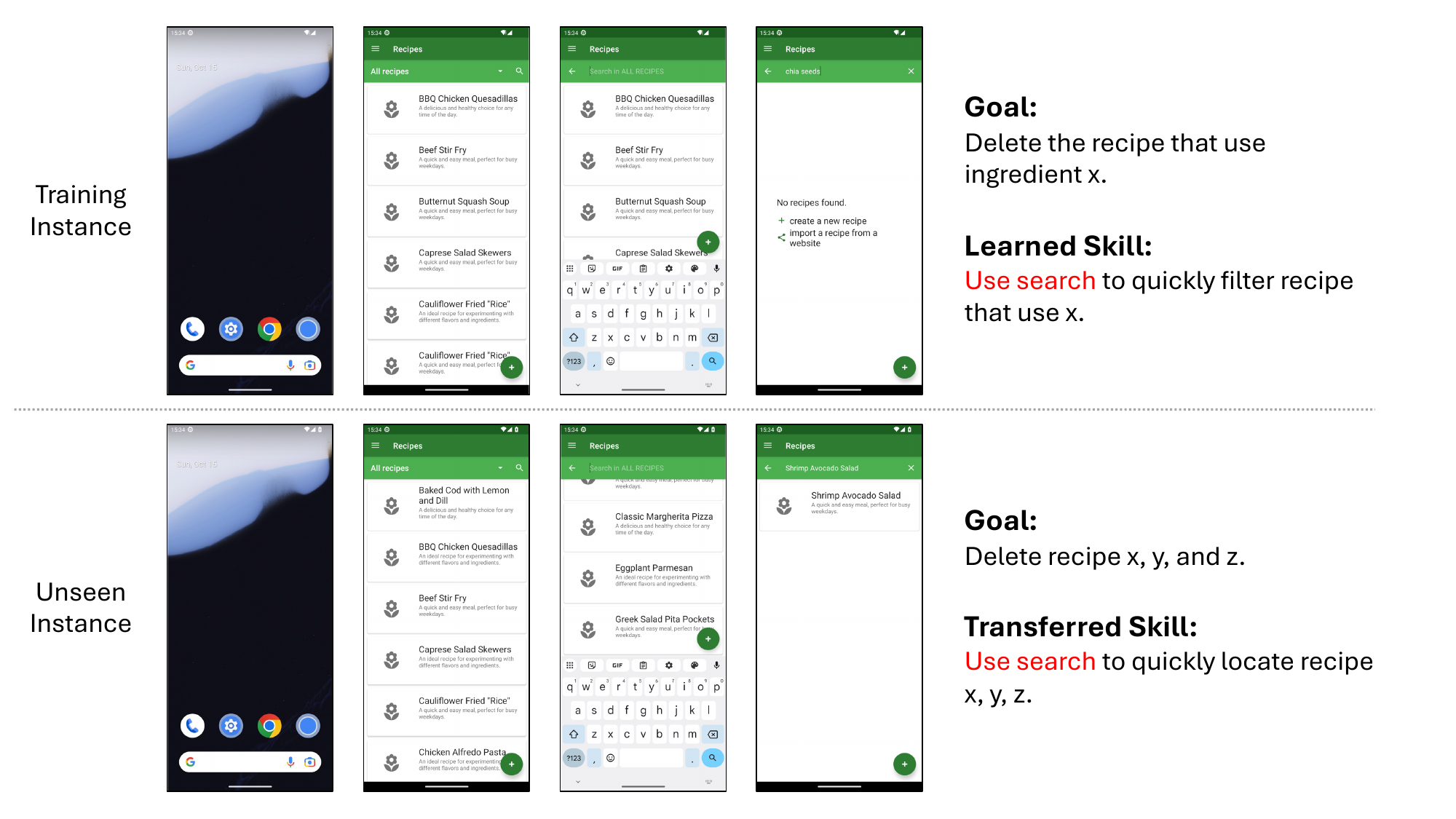}
\caption{Case study 1.}
\label{fig:case_study_1}
\end{figure*}

\begin{figure*}[h!]
\centering
\includegraphics[width=\textwidth]{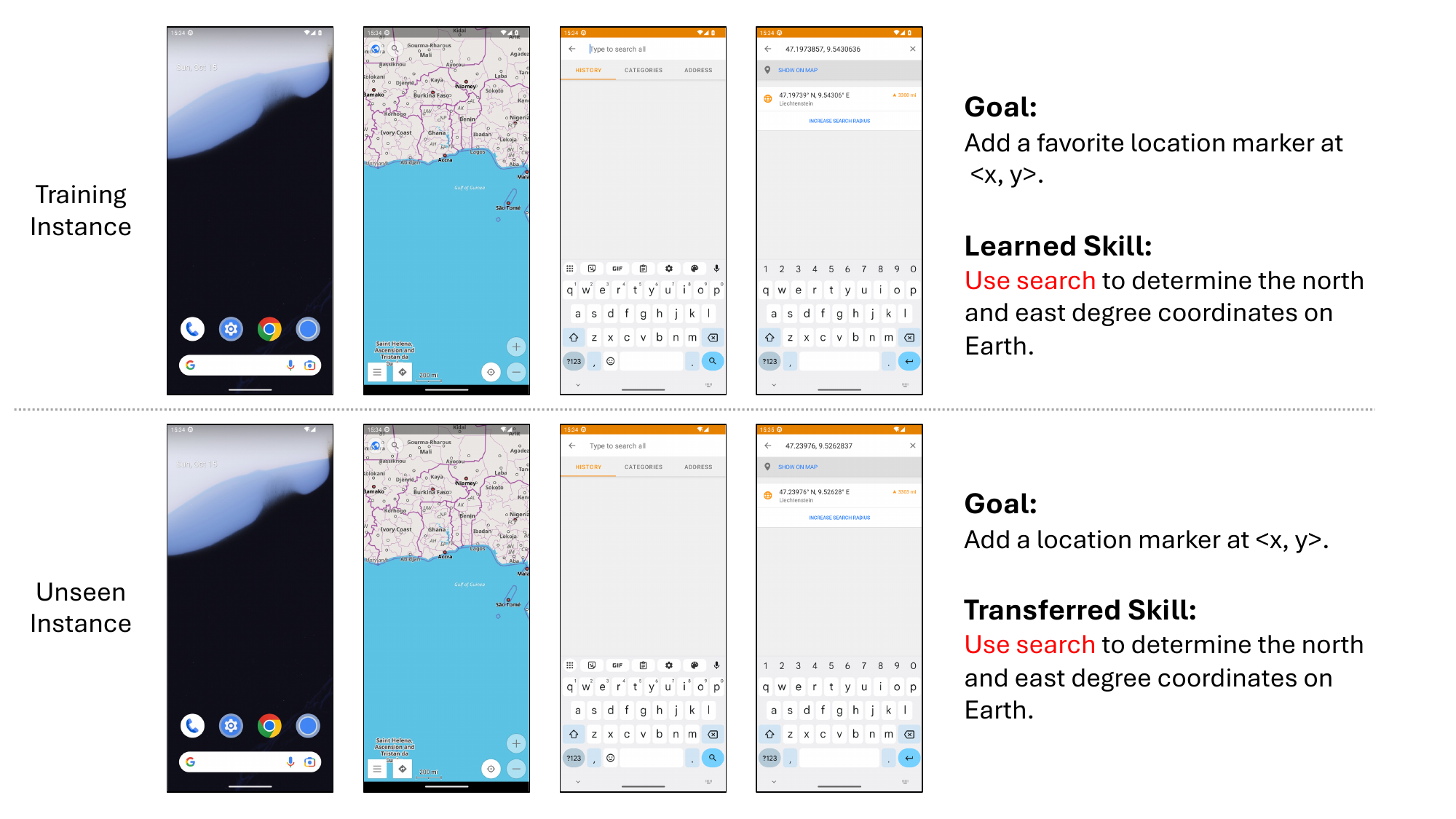}
\caption{Case study 2.}
\label{fig:case_study_2}
\end{figure*}

\begin{figure*}[h!]
\centering
\includegraphics[width=\textwidth]{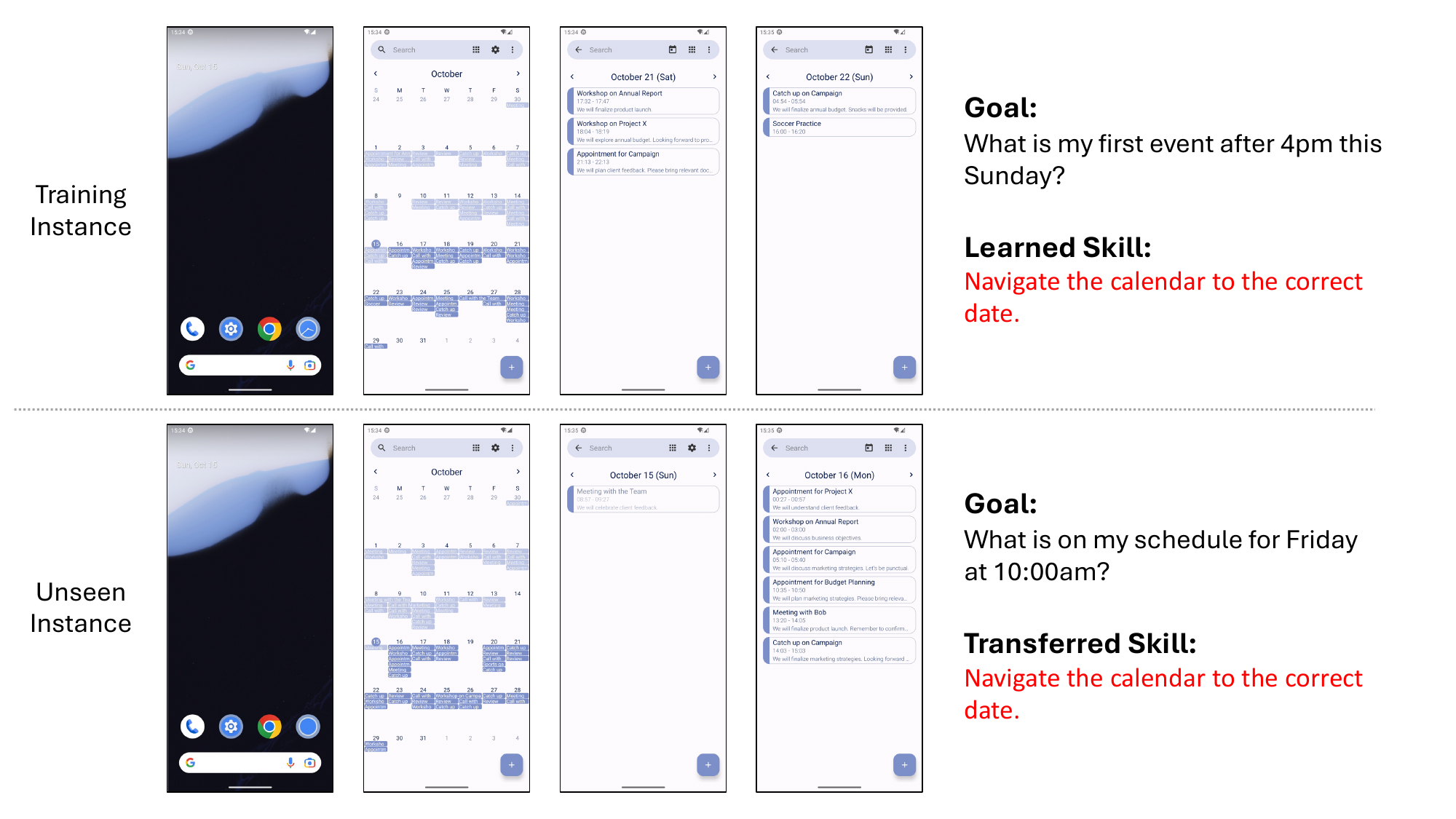}
\caption{Case study 3.}
\label{fig:case_study_3}
\end{figure*}

\begin{figure*}[h!]
\centering
\includegraphics[width=\textwidth]{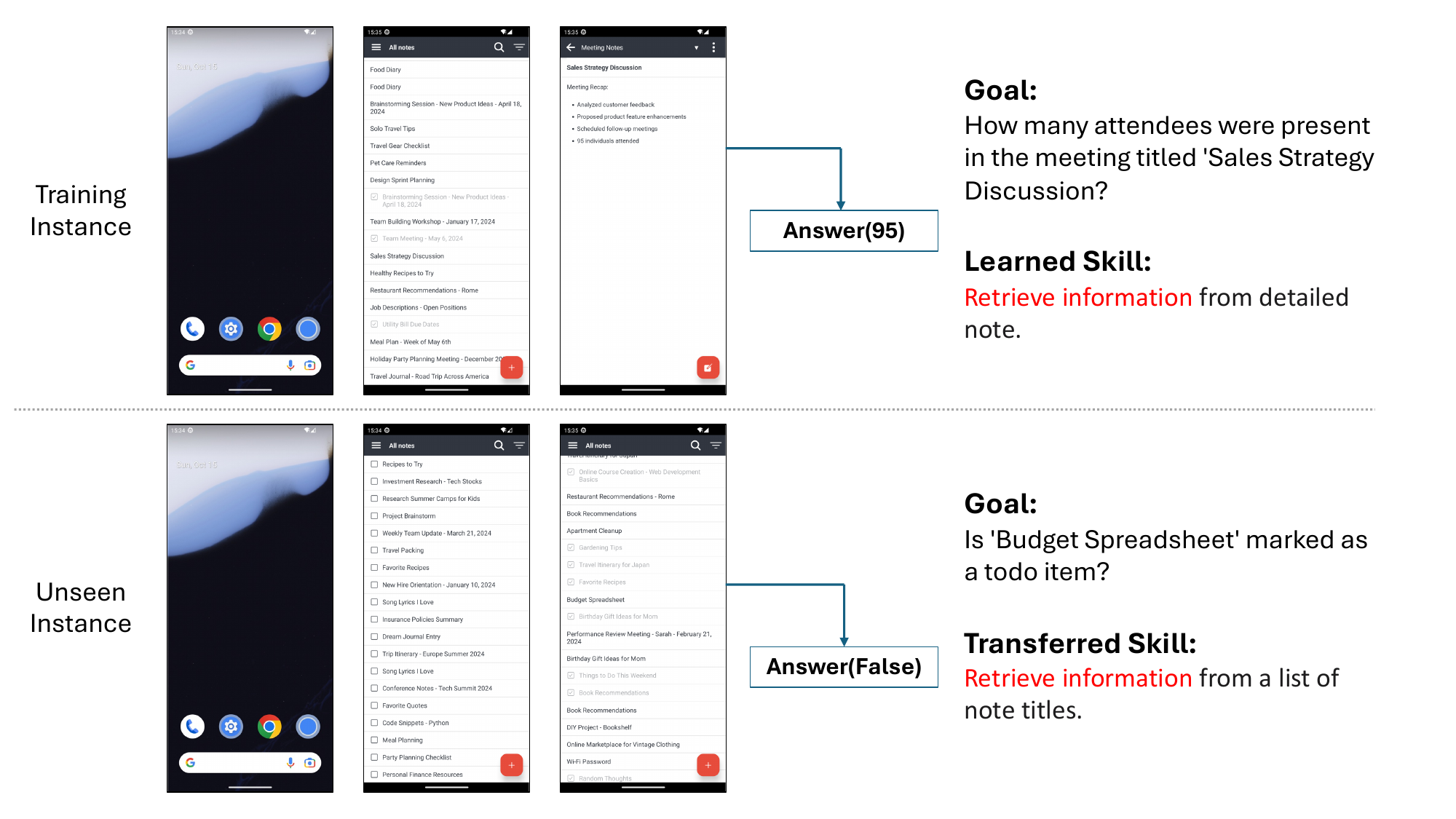}
\caption{Case study 4.}
\label{fig:case_study_4}
\end{figure*}

\begin{figure*}[h!]
\centering
\includegraphics[width=\textwidth]{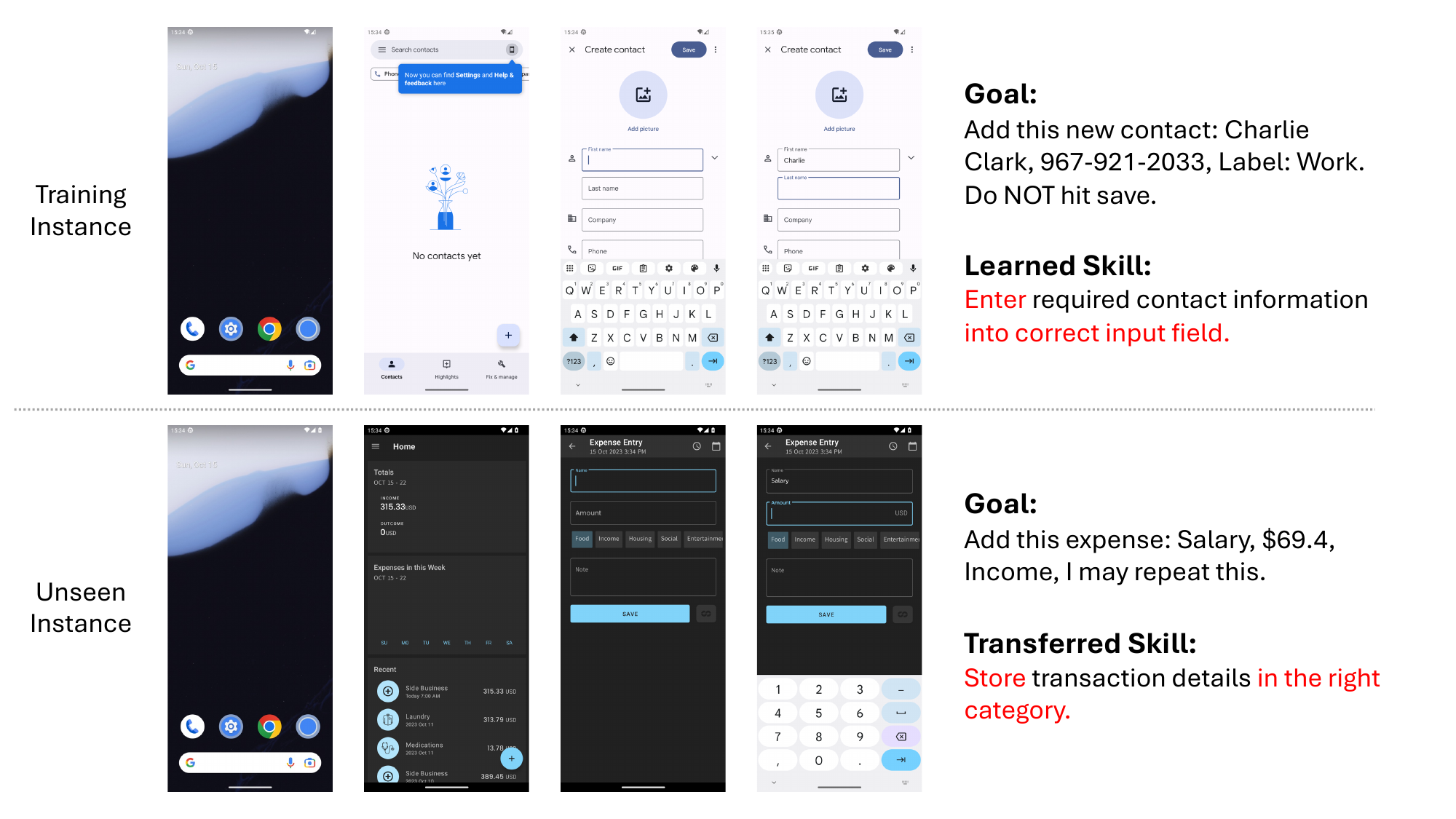}
\caption{Case study 5.}
\label{fig:case_study_5}
\end{figure*}





\section{Few-Shot Adaptation at Test Time}\label{appendix:fewshot}

In the Unseen App regime, the test set contains 16 task templates across 5 unseen
applications: Audio Recorder (1 template), Clock (1), OSMAnd (2), Tasks (4), and
Broccoli (8). To enable few-shot adaptation at test time, we construct a
corresponding training set from these templates. Specifically, we sample 8, 8, 4, 2,
and 1 task instances from the 1, 1, 2, 4, and 8 templates respectively, using
different random seeds. This yields a balanced training set, denoted as
\texttt{unseen-app-train}, in which each unseen application contains 8 distinct task
instances. We manually verified that none of these instances overlap with those in
the test set.

Starting from a model trained for 500 steps on seen apps as the non-adaptation
baseline, we allow 50 additional fine-tuning steps on \texttt{unseen-app-train} at
test time using the rule-based reward function, as the adaptation stage in practical
deployment typically operates on resource-constrained devices and must rely on
few-shot data and limited computation.

We compare two adaptation strategies:
\begin{itemize}
    \item \textbf{All-App}: The model is fine-tuned on all
    \texttt{unseen-app-train} instances jointly across all 5 unseen applications.
    \item \textbf{Per-App}: Separate models are fine-tuned on the 8 instances of
    each unseen application independently, enabling stronger personalization.
\end{itemize}

As shown in \autoref{fig:unseen_app_tta_v2}, Per-App adaptation outperforms the
non-adapted baseline by 22.9 and All-App adaptation yields a transient gain peaking at 8.3, highlighting that
app-specific fine-tuning is more effective than joint adaptation across all unseen
applications.

\section{Ablation study on Rollout Collection System}
\label{rollout_ablate}

To highlight the importance of parallel rollout collection and system optimization, we profile the time of a single policy update iteration, which consists of a policy parameter update and the collection of 16 rollouts executed sequentially and in parallel. For this analysis, we select a task instance with 15 steps, reflecting the average length of training tasks, to generate the rollouts, and repeat the experiment three times for robustness. To ensure a fair comparison, we restart the server before each run and ensure that no other processes occupy GPU or CPU resources. The results show that the update time of the policy parameters remains identical between both settings, while rollout collection occupies more than 95\% of the total runtime in the sequential case and almost 75\% in the parallel case. These findings demonstrate that optimizing rollout collection is essential to alleviate the training bottleneck and accelerate the overall process.

In our asynchronous design, we adopt a first-come, first-served strategy: the VLM generates an action as soon as any environment returns its observation, rather than waiting for all environments to complete a full batch at each rollout step. This allows more actions to be executed concurrently and reduces GPU idle time. To evaluate the effectiveness of this design, we ablate the asynchronous mechanism by disabling it and profile the runtime required to complete all 116 task instances in original AndroidWorld evaluation. Varying the number of environments from 4 to 16, we observe that the asynchronous design yields greater benefits as the number of environments increases, since synchronous models are increasingly delayed by the slowest straggler.

\section{Reward Functions: Rule-Based Scripts vs. LLM-as-Judge}
\label{llm_judge}
To assess the reliability of reward functions, we compare rule-based scripts with an LLM-as-judge setup by training the same mobile agent using Gemini-2.5-Pro as the reward provider. All training hyper-parameters are kept identical, with the only change being the trajectory reward source (rule-based vs. Gemini). During training, Gemini receives the task description and interaction history (screenshots and actions) as input and produces a binary success score via chain-of-thought reasoning. To mitigate hallucinations, the agent’s internal thoughts are excluded from the interaction history. The full prompt is provided below.

\begin{lstlisting}[style=prompt]
SYSTEM_PROMPT = """You are an expert evaluator. Your job is 
to determine whether the assigned task has been successfully 
completed. Base your judgment strictly on visible, verifiable 
evidence from the screenshots and action history. Be concise, 
objective, and avoid making assumptions beyond what is shown."""

COT_PROMPT = """Task: {task}

Respond using this format:
Thinking: <your concise thought and reasoning process>.
Status: success or failure

Guidelines:
- Include only the two fields above.
- For information retrieval tasks (typically where 
  the task requires the agent to provide an answer), 
  confirm that the content in the 'answer' action is 
  accurate and visibly supported by the screenshots.
- For all other tasks, verify that the task has 
  been completed based on clear and observable evidence."""
\end{lstlisting}

\begin{figure*}[h!]
\centering
\begin{minipage}[t]{0.48\linewidth}
    \centering
    \includegraphics[width=\linewidth]{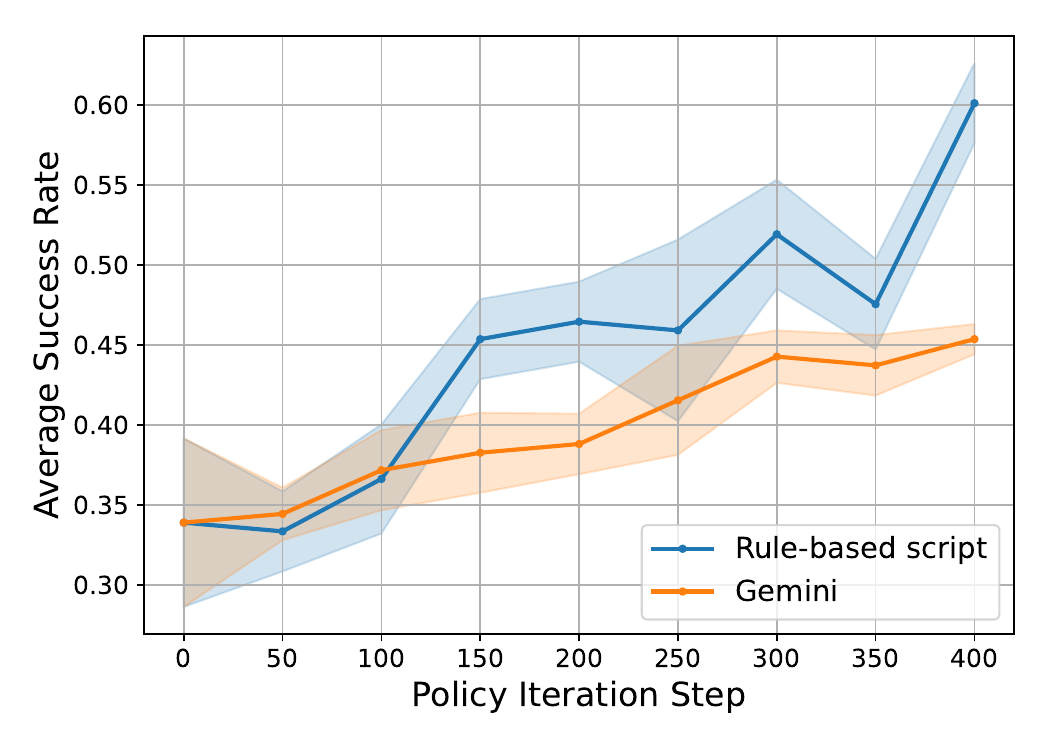}
    \caption{Evaluation success rates using a rule-based script versus Gemini as reward functions. The rule-based script proves more effective at the same number of policy iteration steps.}
    \label{fig:gemini_vs_gt_agent_evaluation_curve}
\end{minipage}
\hfill
\begin{minipage}[t]{0.48\linewidth}
    \centering
    \includegraphics[width=\linewidth]{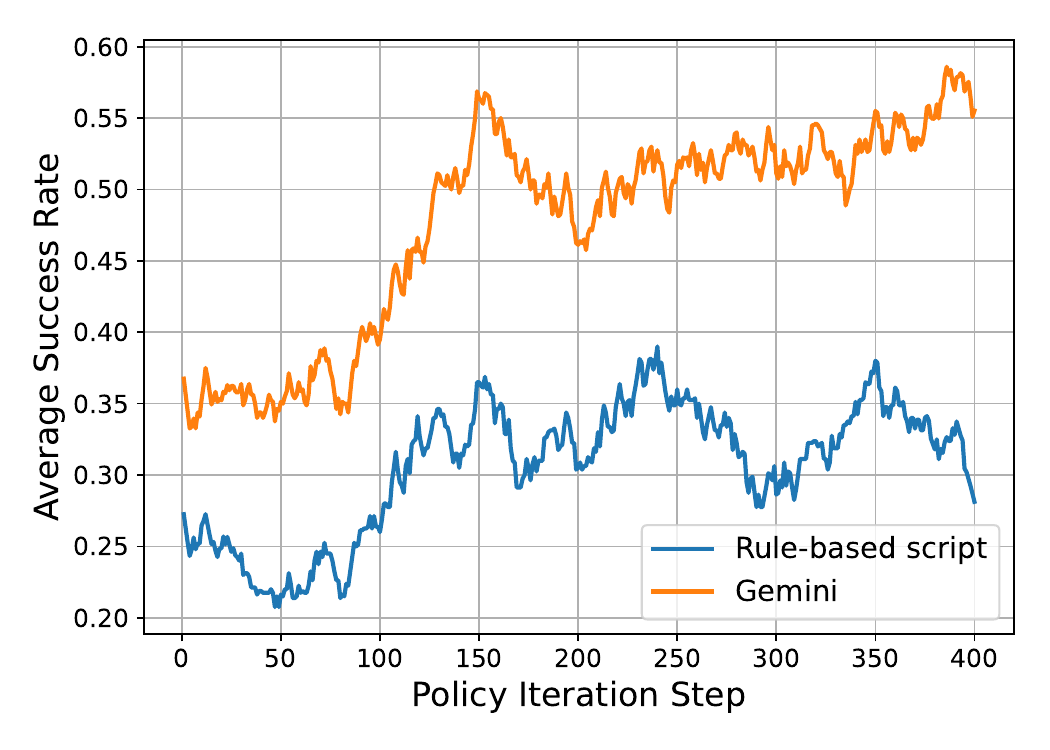}
    \caption{Training success rates using a rule-based script versus Gemini as reward functions. The rule-based script provides reliable rewards, whereas Gemini assigns higher rewards to the same trajectories, leading to false positives and potentially misleading optimization.}
    \label{fig:gemini_vs_gt_training_curve}
\end{minipage}
\end{figure*}

To simplify the experiment, we use the Easy task subset of the Unseen Instance regime, consisting of 553 task instances across 38 templates and 14 apps. Despite differing reward functions during training, evaluation is always performed with the same rule-based script. As shown in \autoref{fig:gemini_vs_gt_agent_evaluation_curve}, the agent trained with the rule-based script achieves a 26.22\% increase in success rate after 400 policy iterations, whereas the agent trained with Gemini reward achieves only an 11.48\% improvement, primarily due to false positive signals produced by Gemini. Specifically, \autoref{fig:gemini_vs_gt_training_curve} shows that Gemini’s reward predictions are consistently higher than the rule-based script's reward feedback, with the gap widening from 10\% to 20\% as training progresses. Consequently, the model is optimized in a misleading direction, highlighting the importance of reliable rule-based reward functions.

\section{Agent Design}
\label{agent_design}
\paragraph{Action space. }The mobile agent interacts with the environment through a set of actions, summarized in \autoref{tab:action_space}. Together, they form a flexible action space that supports both task execution (e.g., completing forms, navigating menus) and information retrieval (e.g., entering queries). This design balances simplicity with expressiveness, ensuring that the agent can operate across diverse applications, task templates and task instances in AndroidWorld.

\paragraph{Prompt Template.} 
We employ a VLM as the policy model to generate \textit{thoughts} followed by a predicted \textit{action}, conditioned on the current screenshot and the interaction history. 
To standardize the model’s input–output format, we adopt a structured prompt template that specifies how task instructions, screenshots, and histories are provided as input, and how the model should output its reasoning process and corresponding action, as follows.

\begin{lstlisting}[style=prompt]
UITARS_USR_PROMPT_THOUGHT = """You are a GUI agent. You are given a task and your action history, with screenshots. You need to perform the next action to complete the task. 

## Output Format
```
Thought: ...
Action: ...
```

## Action Space
click(start_box='<|box_start|>(x1,y1)<|box_end|>')
long_press(start_box='<|box_start|>(x1,y1)<|box_end|>', time='')
type(content='') # If you want to submit your input, use "\\n" at the end of `content`.
scroll(start_box='<|box_start|>(x1,y1)<|box_end|>', end_box='<|box_start|>(x3,y3)<|box_end|>')
press_home()
press_back()
open_app(content='') # Open an app specified by `content`.
finished(content='') # Submit the task regardless of whether it succeeds or fails.
answer(content='') # Answer user's question.

## Note
- Use English in `Thought` and `Action` part.
- Write a small plan and finally summarize your next action (with its target element) in one sentence in `Thought` part.

## User Instruction
{instruction}
"""
\end{lstlisting}

\begin{table*}[h!]
\centering
\begin{tabular}{ll}
\toprule
\textbf{Action Type} & \textbf{Description} \\
\midrule
Click $(x, y)$   & Tap at screen coordinate $(x, y)$ \\
Long press $(x, y)$   & Long press at screen coordinate $(x, y)$ \\
Type $(\text{text})$ & Enter variable-length natural-language text \\
Scroll $(d)$      & Scroll in direction $d \in \{\text{up, down, left, right}\}$ \\
Navigate home $()$ & Navigate back to home screen \\
Navigate back $()$ & Navigate back to previous screen \\
Open $(\text{app})$ & Launch the specified application by name \\
Finish $(\text{text})$ & Report termination of the task \\
Answer $(\text{text})$ & Answer user's question (for Information Retrieval Tasks) \\
\bottomrule
\end{tabular}
\caption{Action space of mobile use agents in AndroidWorld.}
\label{tab:action_space}
\end{table*}

\begin{table*}[t]
\centering
\small
\renewcommand{\arraystretch}{1.2}
\begin{tabular*}{\textwidth}{@{\extracolsep{\fill}}llcccc@{}}
\toprule
    &  & \multicolumn{2}{c}{Success rate (\%)} & \multicolumn{2}{c}{Gain (points)} \\
\cmidrule(lr){3-4} \cmidrule(l){5-6}
Model & Regime ($T$) & Base & Post-RL & Bootstrap 95\% CI & $t$ 95\% CI \\
\midrule
\multirow{3}{*}{Qwen3-VL}
    & Unseen Instance (78) & $51.7\;[42.3,\,61.1]$ & $62.0\;[52.1,\,71.4]$ & $+10.3\;[+4.3,\,+16.7]$ & $[+3.7,\,+16.8]$ \\
    & Unseen Template (18) & $44.4\;[25.9,\,63.0]$ & $51.9\;[31.5,\,72.2]$ & $+7.4\;[+1.9,\,+14.8]$  & $[+0.3,\,+14.5]$ \\
    & Unseen App (16)      & $41.7\;[20.8,\,64.6]$ & $41.7\;[20.8,\,64.6]$ & $+0.0\;[-12.5,\,+12.5]$ & $[-14.5,\,+14.5]$ \\
\midrule
\multirow{3}{*}{UI-TARS}
    & Unseen Instance (78) & $18.8\;[11.1,\,27.4]$ & $36.3\;[27.4,\,45.3]$ & $+17.5\;[+9.4,\,+25.6]$ & $[+9.3,\,+25.7]$ \\
    & Unseen Template (18) & $20.4\;[5.6,\,38.9]$  & $31.5\;[13.0,\,51.9]$ & $+11.1\;[+3.7,\,+18.5]$ & $[+3.1,\,+19.2]$ \\
    & Unseen App (16)      & $22.9\;[8.3,\,41.7]$  & $27.1\;[10.4,\,45.8]$ & $+4.2\;[+0.0,\,+10.4]$  & $[-1.9,\,+10.2]$ \\
\bottomrule
\end{tabular*}
\caption{Success rates and paired per-template gains for the Q2 runs (500 steps, no
curriculum), with 95\% percentile-bootstrap confidence intervals over templates and the
Student-$t$ interval on the gain as a cross-check.}
\label{tab:gain_ci}
\end{table*}

\section{Interactive Environment}
\label{environment}
We follow the AndroidWorld setup procedure to construct our mobile emulator environment on a headless Linux server. The emulator is configured with Android API level 33 (Tiramisu), a resolution of 2400×1080, 16 GB RAM, 6 GB disk space, and 6 virtual CPU cores. All required apps are preinstalled, and their initial launch states are manually verified for correctness. A clean emulator snapshot is saved to eliminate repeated setup overhead during parallel execution. To accommodate action execution, we enforce a fixed delay of 3 seconds and extend it dynamically up to 6 seconds until screen stabilization, determined by comparing accessibility trees across consecutive frames. In contrast, the communication overhead between the Docker server and agents is negligible relative to action latency.

\section{Training Details}
\label{implementation}
We use \textbf{UI-TARS-7B-SFT} as the base model. The mobile agent prompt is adapted from the original UI-TARS codebase with minimal modification, adding the required \textit{answer} action to handle \textit{information retrieval} tasks. At each training step, 2 task instances are sampled, and 8 rollout trajectories are collected per instance. Full trajectories are used for weight updates, with trajectory length capped at 20 steps in addition to AndroidWorld task-specific limits to accommodate GPU memory constraints. Under this configuration, each experiment uses 16 virtual environments for trajectory collection and 4 NVIDIA H100 80GB GPUs for action generation and weight updates. All experiments are conducted on a 224-core server with 8 H100 GPUs, supporting up to 2 concurrent training runs (32 Docker containers across 8 GPUs) without latency degradation; performance deteriorates significantly beyond this scale. Training is performed with a maximum learning rate of $1\times10^{-6}$, linearly warmed up over 90 steps, and a KL-divergence coefficient of 0.05. Input images are resized to 1120×504, a sampling temperature of 1 is applied, and the main experiment runs for 2000 training steps, requiring approximately 700 GPU hours.


\section{Confidence Intervals for Per-Regime Gains}
\label{appendix:ci}


\paragraph{Setup.}
This appendix reports intervals for the Q2 runs: a 500-step budget without curriculum,
with every other setting held constant across regimes and priors. The unit of analysis is
the task template in all three unseen regimes. Each held-out set instantiates its
templates with three independent parameter draws, so instances are correlated
measurements of a template rather than independent trials: the held-out sets of
$n=234/54/48$ instances correspond to $T=78/18/16$ templates, and $T$ is the effective
sample size for every interval below. The prior (step~0) and post-RL (step~500)
checkpoints are evaluated on identical instances, giving a paired comparison in which
each template's gain is the difference of its two success rates, each averaged over the
three draws. \autoref{tab:gain_ci} reports 95\% percentile-bootstrap confidence intervals
over templates (10{,}000 resamples, seed~0) for the gains and the marginal success rates,
with Student-$t$ intervals on the gains as a cross-check. Intervals cover the GRPO runs;
the PPO replication in \autoref{fig:ppo_unseen_scenarios} is reported as point estimates. Two cautions
apply: the success-rate intervals are marginal, so their overlap does not contradict a
nonzero paired gain; and entries are rounded after computation, so Post-RL $-$ Prior can
differ from the tabulated gain by $0.1$, as in the Unseen Template row for Qwen3-VL-8B.
All six gain intervals are reported without multiplicity adjustment and read as
descriptive.

\paragraph{Bootstrap and $t$ intervals.}
The two methods agree everywhere except Unseen App for UI-TARS-7B-SFT. With $T=16$
templates and three draws each, per-template gains fall on a seven-point lattice from
$-1$ to $+1$ in thirds; the percentile endpoint lands on an atom of that lattice, which
is what produces a lower bound of exactly $+0.0$. We read neither Unseen App interval as
resolving the sign of the gain, and note that UI-TARS-7B-SFT's $+4.2$ corresponds to two
instances out of $48$.


\paragraph{Reading the intervals.}
Point estimates decay monotonically across regimes for both priors, matching
\autoref{fig:ablation_study_on_unseen_scenarios}. For both, the Unseen Instance and Unseen Template gains exclude zero,
whereas neither Unseen App gain is resolvably positive: Qwen3-VL-8B sits at zero with a
symmetric interval, and UI-TARS-7B-SFT's lower endpoint touches zero. Adjacent regimes'
gain intervals overlap substantially, and because the three regimes draw on disjoint
template populations, adjacent-regime differences are not paired and we do not interval
them. We therefore claim the ordering of regimes and the widening train--test gap rather
than a separable difference between individual regimes.


\end{document}